\documentclass[lettersize,journal]{IEEEtran}
\usepackage{amsmath,amsfonts}
\usepackage{algorithmic}
\usepackage{algorithm}
\usepackage{array}
\usepackage[caption=false]{subfig}
\usepackage{textcomp}
\usepackage{stfloats}
\usepackage{url}
\usepackage{verbatim}
\usepackage{graphicx}
\usepackage{cite}
\usepackage{amsmath}
\usepackage{amssymb}
\usepackage{multirow}
\usepackage{colortbl}
\usepackage{booktabs}
\usepackage{makecell}
\usepackage{bbding}
\usepackage{pifont}
\usepackage{setspace}
\usepackage{threeparttable}
\usepackage{balance}
\usepackage{tikz,xcolor}

\usepackage[colorlinks,
linkcolor=blue,
anchorcolor=blue,
citecolor=blue]{hyperref}

\hypersetup{hidelinks,
	colorlinks=true,
	allcolors=blue,
	pdfstartview=Fit,
	breaklinks=true}

\definecolor{lime}{HTML}{A6CE39}
\DeclareRobustCommand{\orcidicon}{
\begin{tikzpicture}
\draw[lime, fill=lime] (0,0)
circle[radius=0.16]
node[white]{{\fontfamily{qag}\selectfont \tiny \.{I}D}};
\end{tikzpicture}
\hspace{-2mm}
}
\foreach \x in {A, ..., Z}{%
\expandafter\xdef\csname orcid\x\endcsname{\noexpand\href{https://orcid.org/\csname orcidauthor\x\endcsname}{\noexpand\orcidicon}}
}

\begin{document}

%\title{A Sample Article Using IEEEtran.cls\\ for IEEE Journals and Transactions}
\title{Large, Complex, and Realistic Safety Clothing and Helmet Detection: Dataset and Method}
%~\IEEEmembership{~IEEE,}
\author{Fusheng Yu, Jiang Li\hspace{-1.5mm}\orcidA{}, Xiaoping Wang\hspace{-1.5mm}\orcidB{},~\IEEEmembership{Senior Member,~IEEE}, Shaojin Wu, Junjie Zhang, and\\ Zhigang Zeng\hspace{-1.5mm}\orcidC{},~\IEEEmembership{Fellow,~IEEE}
        % <-this % stops a space
        
\thanks{Manuscript received 10 August 2024; revised 10 August 2024; accepted 10 August 2024. This work was supported in part by the Interdisciplinary Research Program of HUST under Grant 2024JCYJ006, the National Natural Science Foundation of China under Grant 62236005 and 61936004, and the Fundamental Research Funds for the Central Universities of HUST under Grant 2024JYCXJJ067 and 2024JYCXJJ068. \textit{(Fusheng Yu and Jiang Li contributed equally to this work. Corresponding author: Jiang Li and Xiaoping Wang.)}}
\thanks{The authors are with the School of Artificial Intelligence and Automation, Huazhong University of Science and Technology, Wuhan 430074, China, also with the Institute of Artificial Intelligence, Huazhong University of Science and Technology, Wuhan 430074, China, also with the Hubei Key Laboratory of Brain-Inspired Intelligent Systems, Huazhong University of Science and Technology, Wuhan 430074, China, and also with the Key Laboratory of Image Processing and Intelligent Control (Huazhong University of Science and Technology), Ministry of Education, Wuhan 430074, China (e-mail: yufusheng@hust.edu.cn; lijfrank@hust.edu.cn; wangxiaoping@hust.edu.cn; wushaojin@hust.edu.cn; zjunjie@hust.edu.cn; zgzeng@hust.edu.cn).}
\thanks{Digital Object Identifier}
}

\markboth{}
{Yu \MakeLowercase{\textit{et al.}}: Large, Complex, and Realistic Safety Clothing and Helmet Detection}

\IEEEpubid{0000--0000~\copyright~2024 IEEE. Personal use is permitted, but republication/redistribution requires IEEE permission.}

\maketitle

\begin{abstract}
Detecting safety clothing and helmets is paramount for ensuring the safety of construction workers. However, the development of deep learning models in this domain has been impeded by the scarcity of high-quality datasets. In this study, we construct a large, complex, and realistic safety clothing and helmet detection (SFCHD) dataset. SFCHD is derived from two authentic chemical plants, comprising 12,373 images, 7 categories, and 50,552 annotations. We partition the SFCHD dataset into training and testing sets with a ratio of 4:1 and validate its utility by applying several classic object detection algorithms. Furthermore, drawing inspiration from spatial and channel attention mechanisms, we design a spatial and channel attention-based low-light enhancement (SCALE) module. SCALE is a plug-and-play component with a high degree of flexibility. Extensive evaluations of the SCALE module on both the ExDark and SFCHD datasets have empirically demonstrated its efficacy in enhancing the performance of detectors under low-light conditions. The dataset and code are publicly available at https://github.com/lijfrank-open/SFCHD-SCALE.
\end{abstract}

\def\abstractname{Note to Practitioners}
\begin{abstract}
This paper was motivated by the challenge of safety clothing and helmet detection in construction safety inspections. At high-risk construction sites, protective gear like safety clothing and helmets is crucial for worker protection. Monitoring their use is of significant safety implication and practical value. Currently, only two helmet detection benchmarks have been published, focusing mainly on helmets and neglecting safety clothing, which could cause safety risks in real applications. Additionally, existing datasets feature overly simplified backgrounds and idealized lighting, differing greatly from actual work conditions. This may lead to poor generalization of models trained on these datasets. To tackle these issues, we introduce a large, complex, and realistic safety clothing and helmet detection (SFCHD) dataset. It presents more authentic work scenarios with intricate backgrounds, with images captured from two chemical plants' 40 different scenes via surveillance cameras. SFCHD offers abundant training samples and benchmarks for various detection tasks, including small object and high-low light detections. We also design the SCALE module to overcome the limitations of existing algorithms under low-light conditions. This plug-and-play module excavates valuable information from low-light images in both spatial and channel dimensions, significantly boosting detection model performance. Our dataset and method can be deployed in real-world construction safety inspections for real-time monitoring of workers' safety gear, ensuring their personal safety.
\end{abstract}


\section{Introduction}
\IEEEPARstart{S}{afety} issues are an enduring and significant concern across all industries, particularly in high-risk construction sites such as chemical plants. Protective equipment, including safety clothing and helmets, is crucial for safeguarding workers in high-risk construction sites. Helmets can effectively prevent head injuries caused by falling or splashing objects, while safety clothing can protect the body and arms from hazardous chemicals and liquids. The absence of safety clothing and helmets frequently results in safety accidents, exerting a devastating impact on families and society. Consequently, monitoring the usage of safety clothing and helmets in factories or construction sites is of immense safety significance and broad application value.

\IEEEpubidadjcol
Currently, the conventional automatic monitoring technology for safety clothing and helmet compliance is a category of sensor-based methods~\cite{Kelm-sensor1,Naticchia-sensor3}. These techniques monitor whether workers are wearing safety attire or helmets by installing sensors on them and analyzing the emitted signals. However, sensor-based approaches necessitate substantial investment in procurement, installation, and maintenance, leading to relatively high costs. In recent years, deep learning methods have gained substantial attention in computer vision~\cite{Hua17, nah12} due to their ability to self-learn useful features from large-scale, annotated training data. In particular, convolution neural networks are widely used for image classification and object detection. For instance, LeCun et al.~\cite{Lecun-4} utilized convolutional neural networks (CNNs) to recognize handwritten digits; Kolar et al.~\cite{Kolar5} applied CNNs to detect safety guardrails in construction sites; Nath et al.~\cite{nath6,nath7} adopted CNNs to identify common construction-related objects, e.g., buildings, equipment, and workers. These endeavors have provided new insights into the monitoring of safety clothing and helmet usage.

The tremendous success of deep learning technique is largely attributed to the availability of large-scale, high-quality, and meticulously annotated datasets. To the best of our knowledge, only two benchmark datasets for helmet detection have been published to date, namely Pictor-v3~\cite{nath8} and SHWD~\cite{gochoo2021safety}. These datasets primarily focus on the detection of helmets, with insufficient consideration given to safety clothing, leading to potential safety hazards in practical applications. Moreover, the image background in existing datasets is overly simplified, and the lighting conditions are too idealized, which is significantly different from real-world work environments. Training models directly on these datasets may lead to insufficient generalization capabilities in their practical applications. Therefore, these two datasets do not fully meet the needs of real-world applications.

\begin{figure*}[htbp]
	\centering
	\subfloat[]{\includegraphics[width=1.3in]{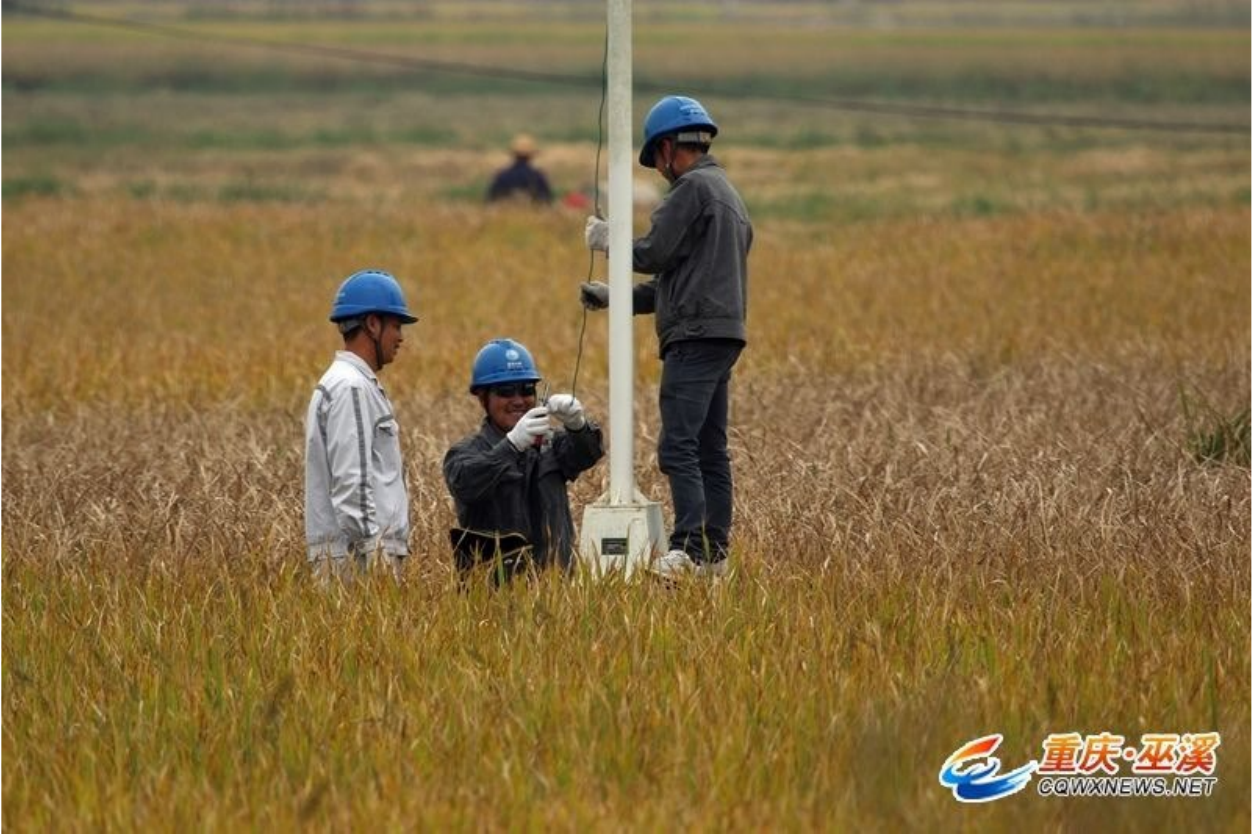}
	\label{fig:pictorv3_1}}
	\hfil
	\subfloat[]{\includegraphics[width=1.3in]{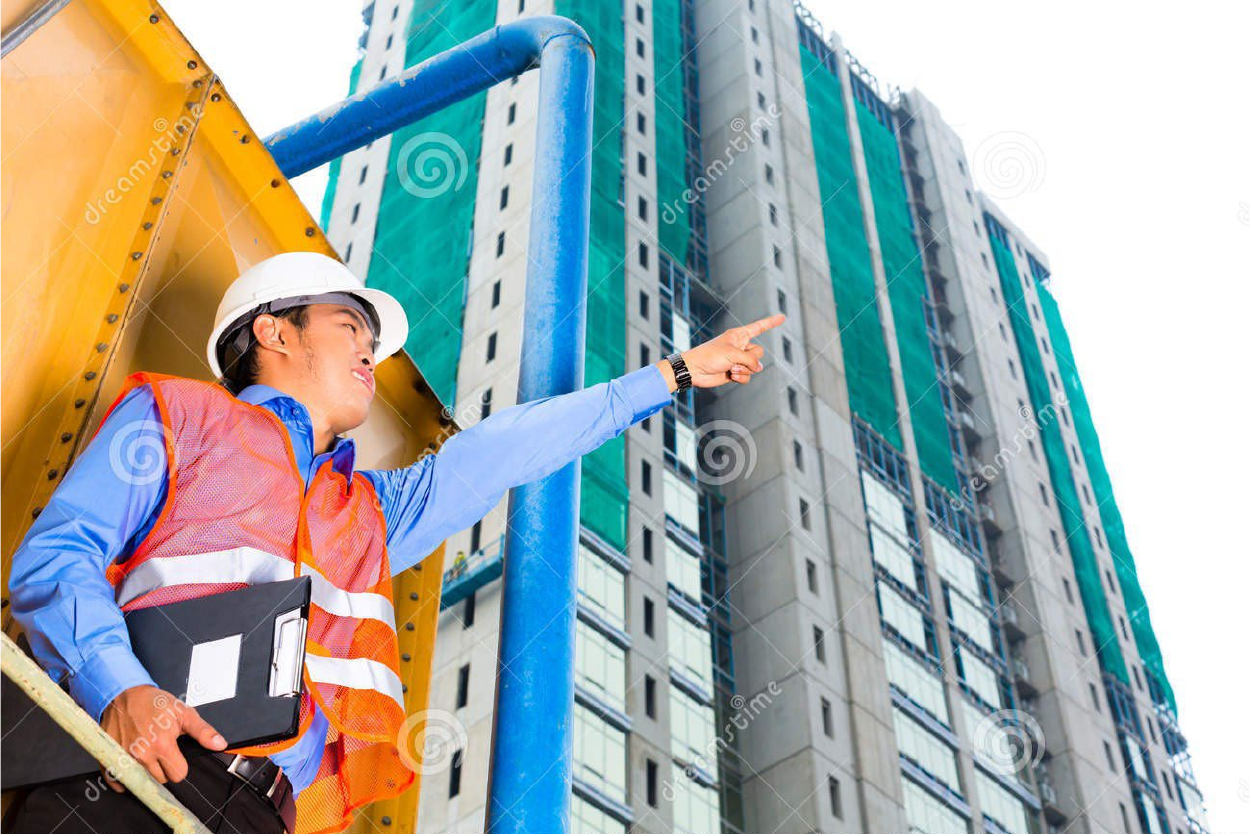}
	\label{fig:pictorv3_2}}
	\hfil
	\subfloat[]{\includegraphics[width=1.3in]{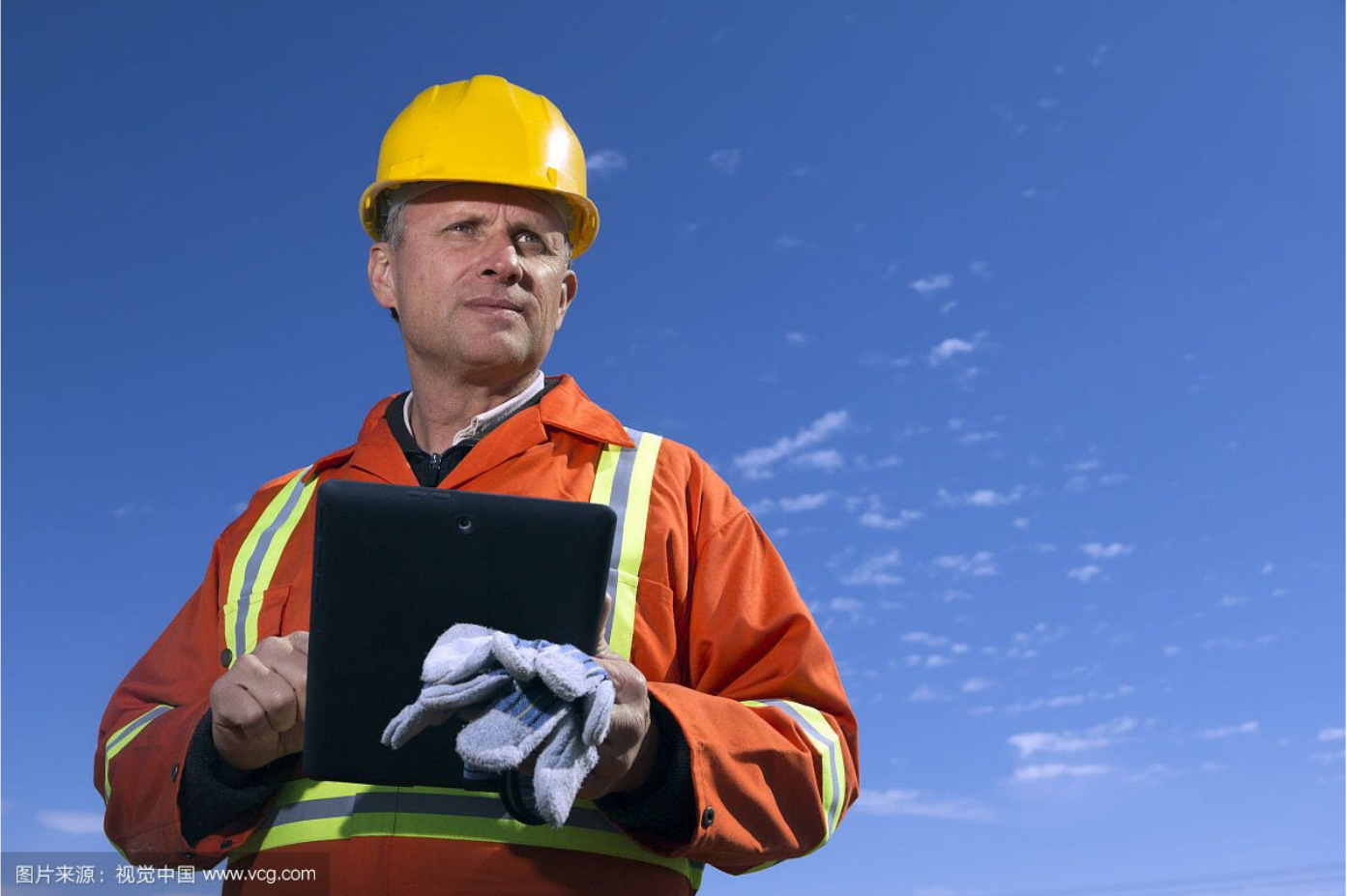}
	\label{fig:pictorv3_3}}
	\hfil
	\subfloat[]{\includegraphics[width=1.3in]{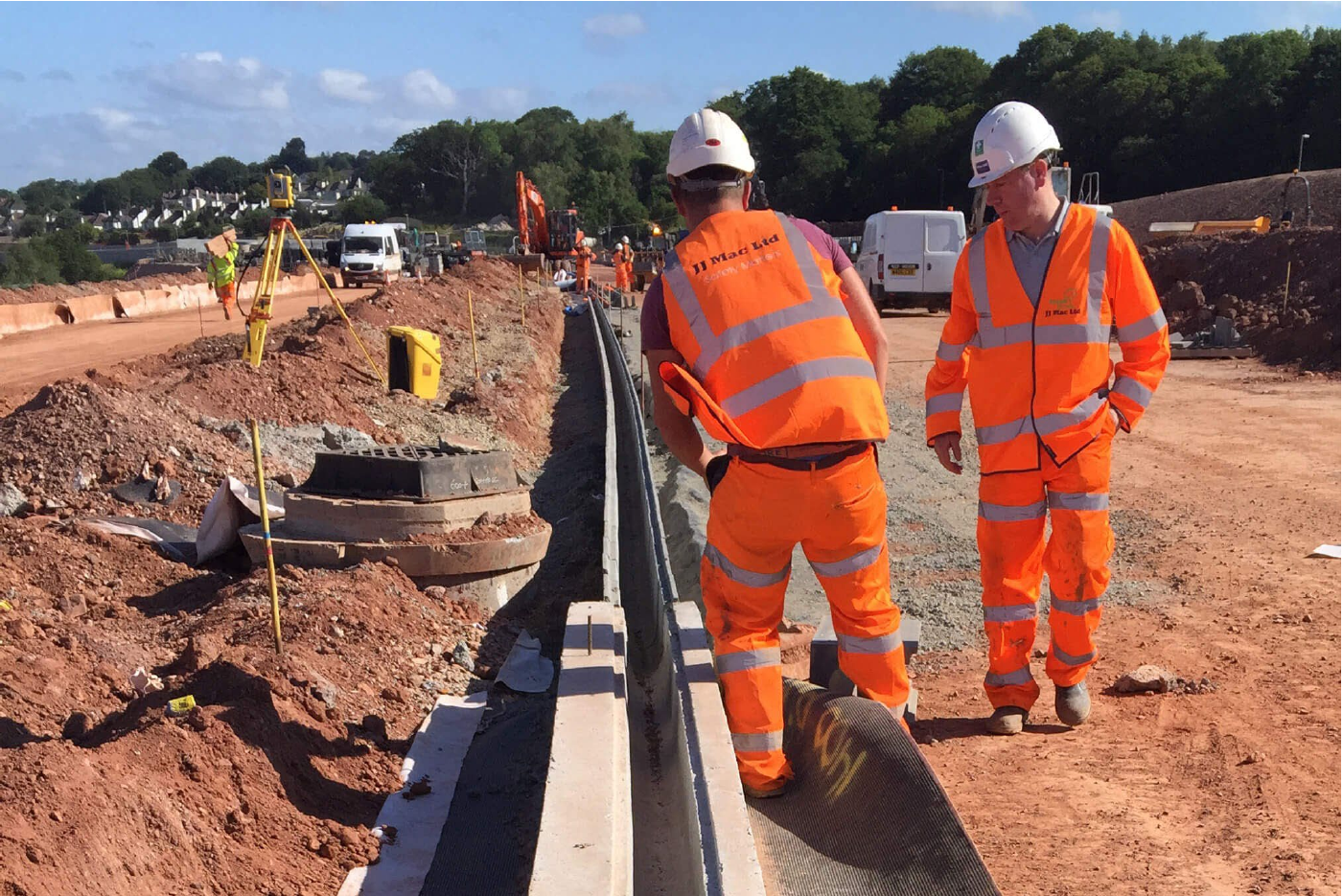}
	\label{fig:pictorv3_4}}
	\hfil
	\subfloat[]{\includegraphics[width=1.3in]{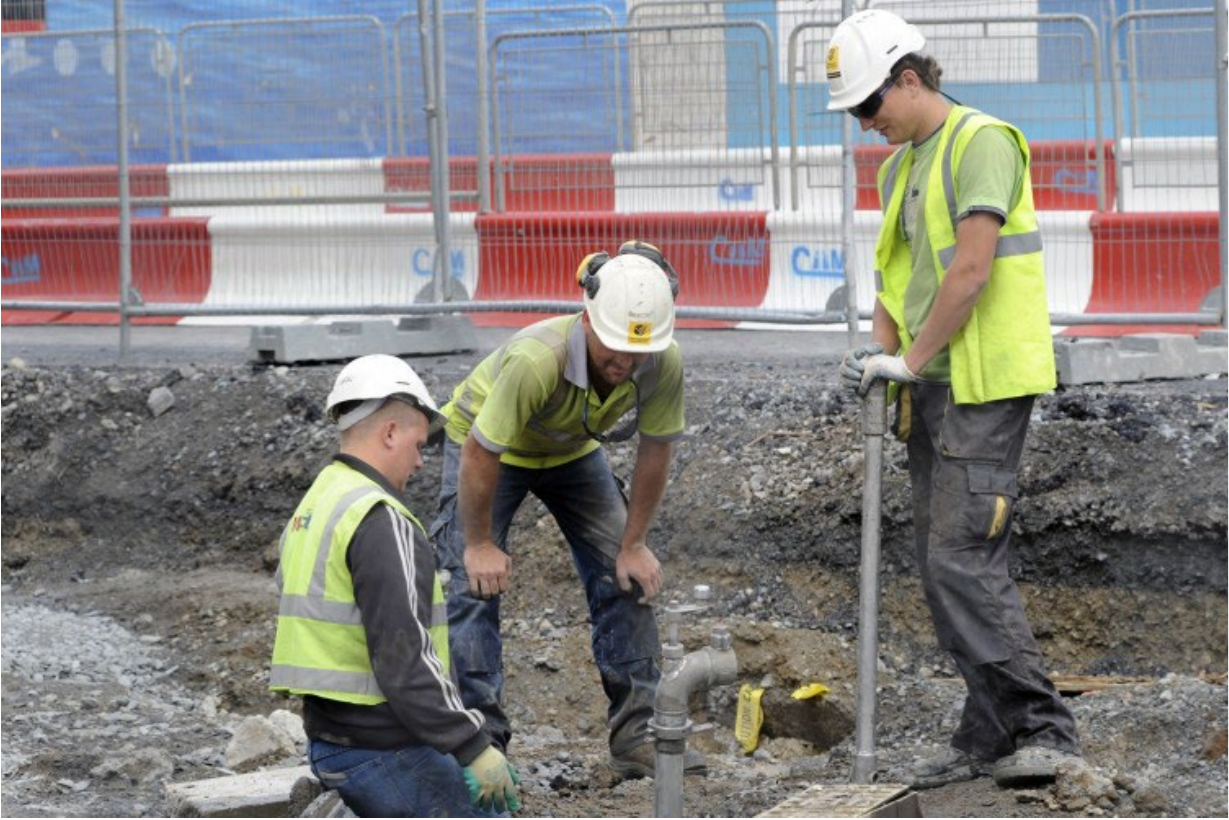}
	\label{fig:pictorv3_5}}
	\vfil
	\subfloat[]{\includegraphics[width=1.3in]{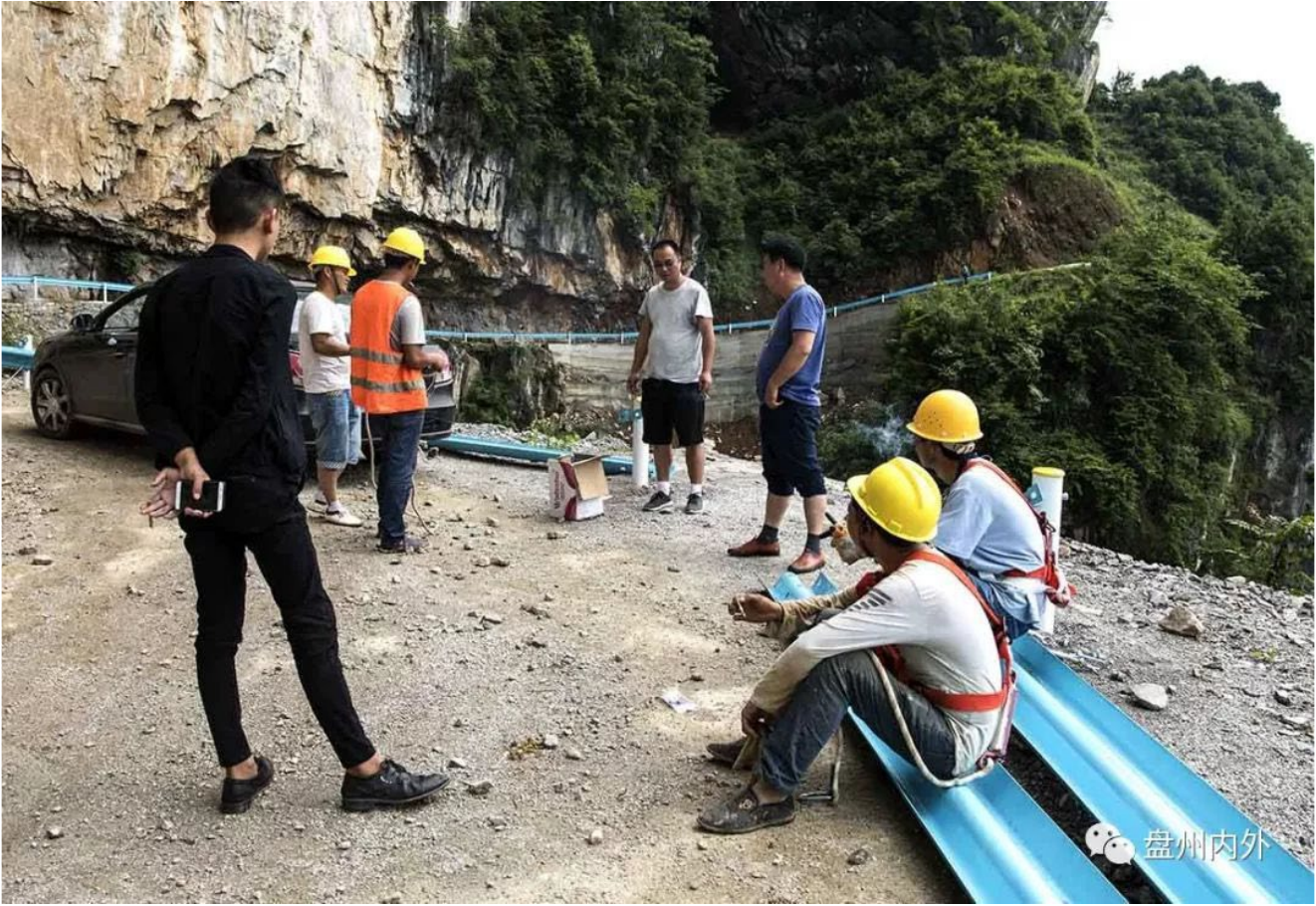}
	\label{fig:shwd_1}}
	\hfil
	\subfloat[]{\includegraphics[width=1.3in]{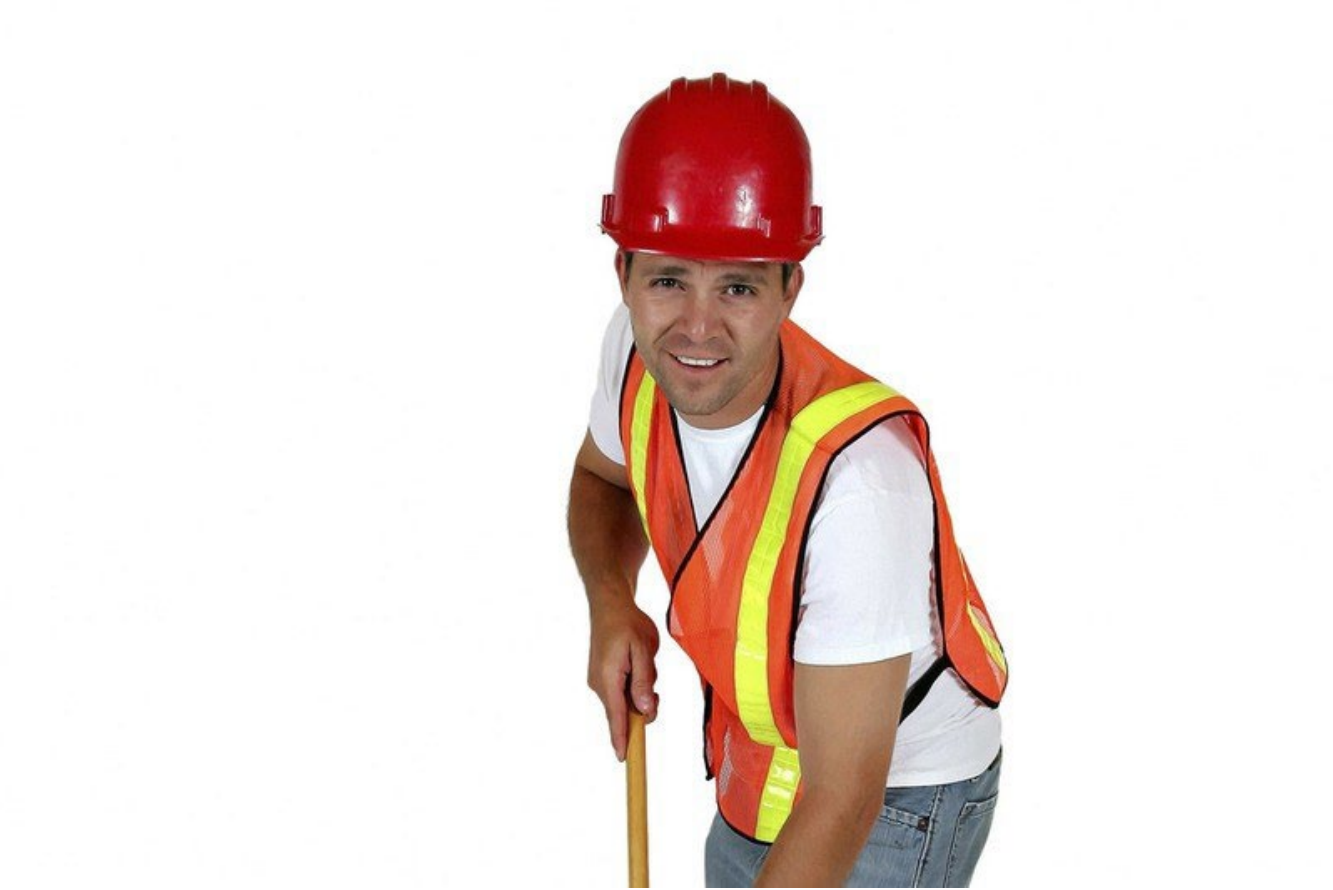}
	\label{fig:shwd_2}}
	\hfil
	\subfloat[]{\includegraphics[width=1.3in]{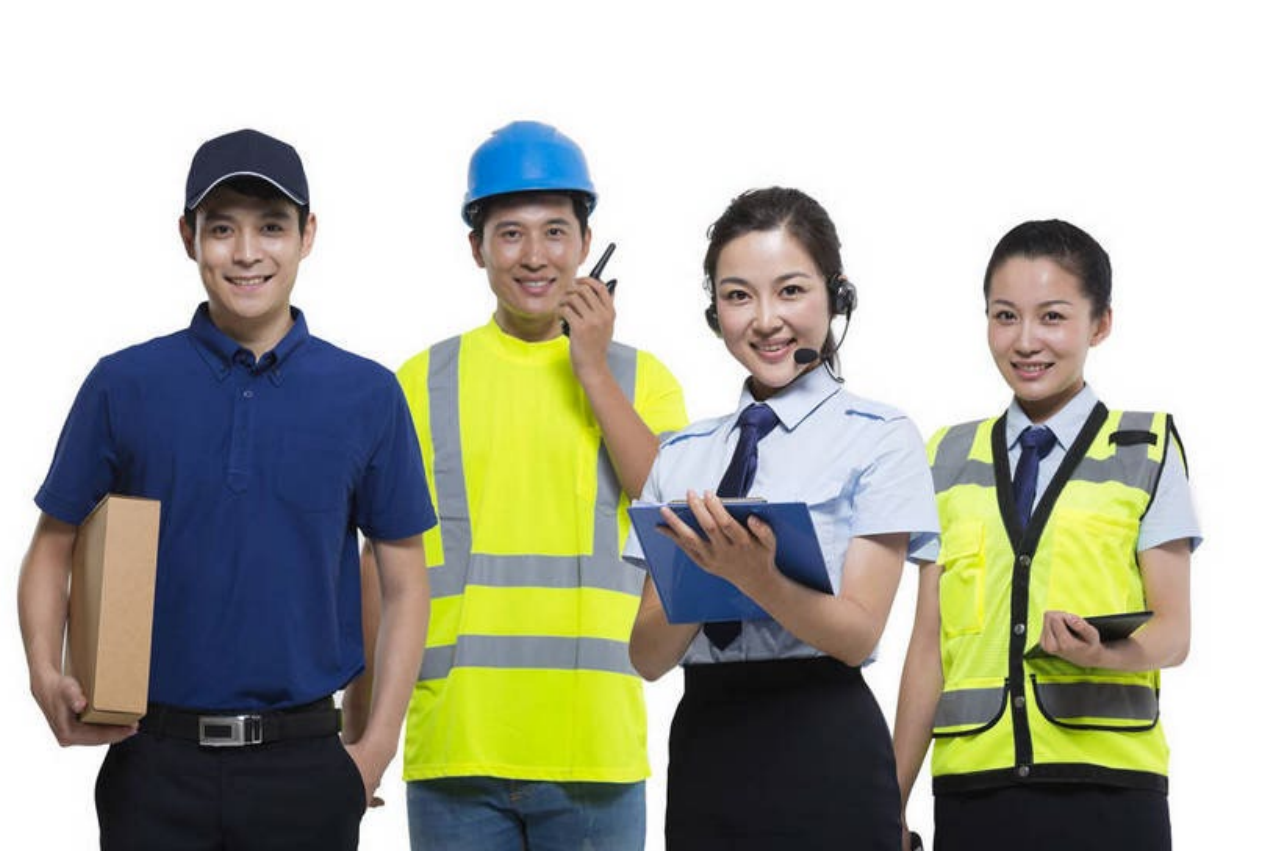}
	\label{fig:shwd_3}}
	\hfil
	\subfloat[]{\includegraphics[width=1.3in]{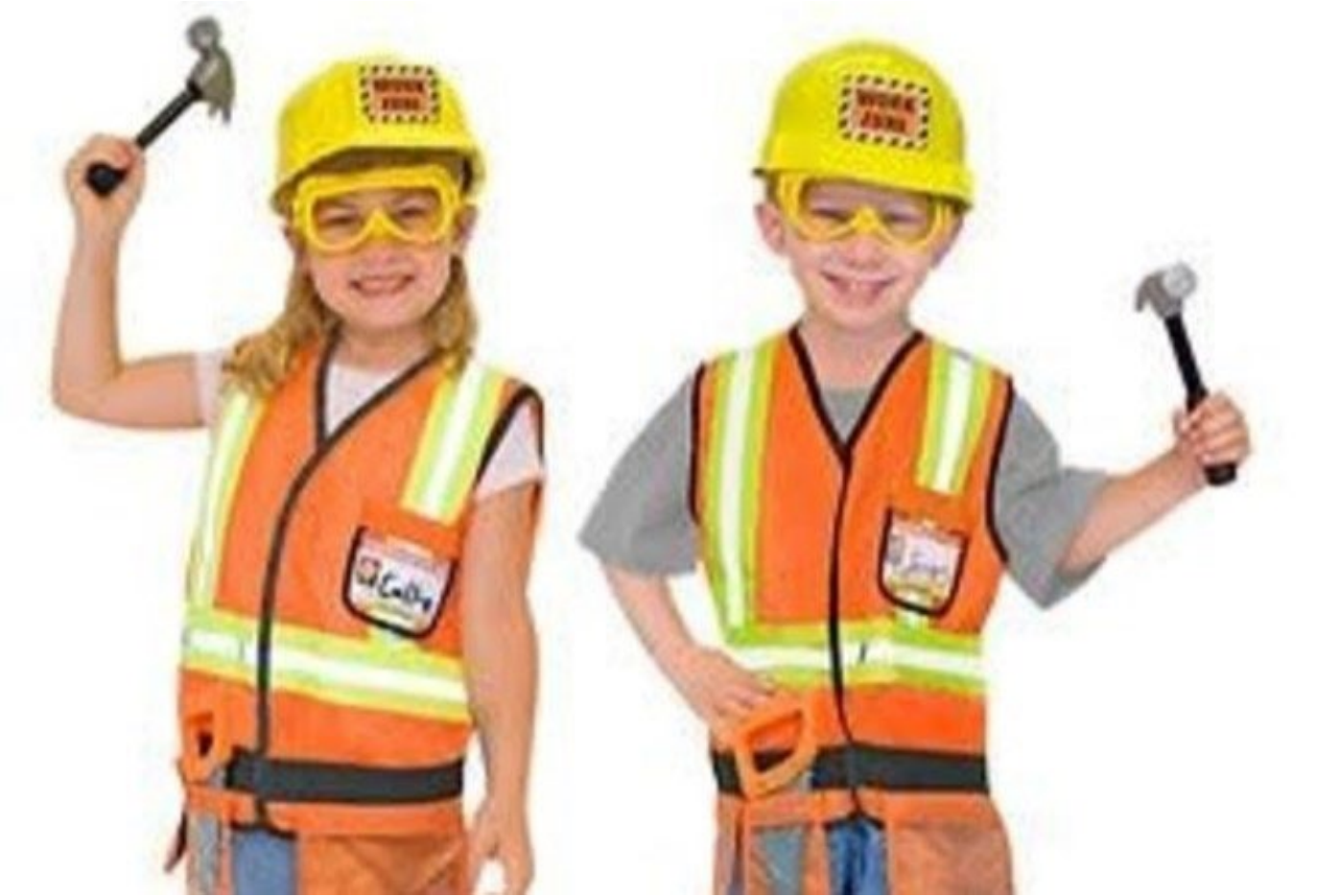}
	\label{fig:shwd_4}}
	\hfil
	\subfloat[]{\includegraphics[width=1.3in]{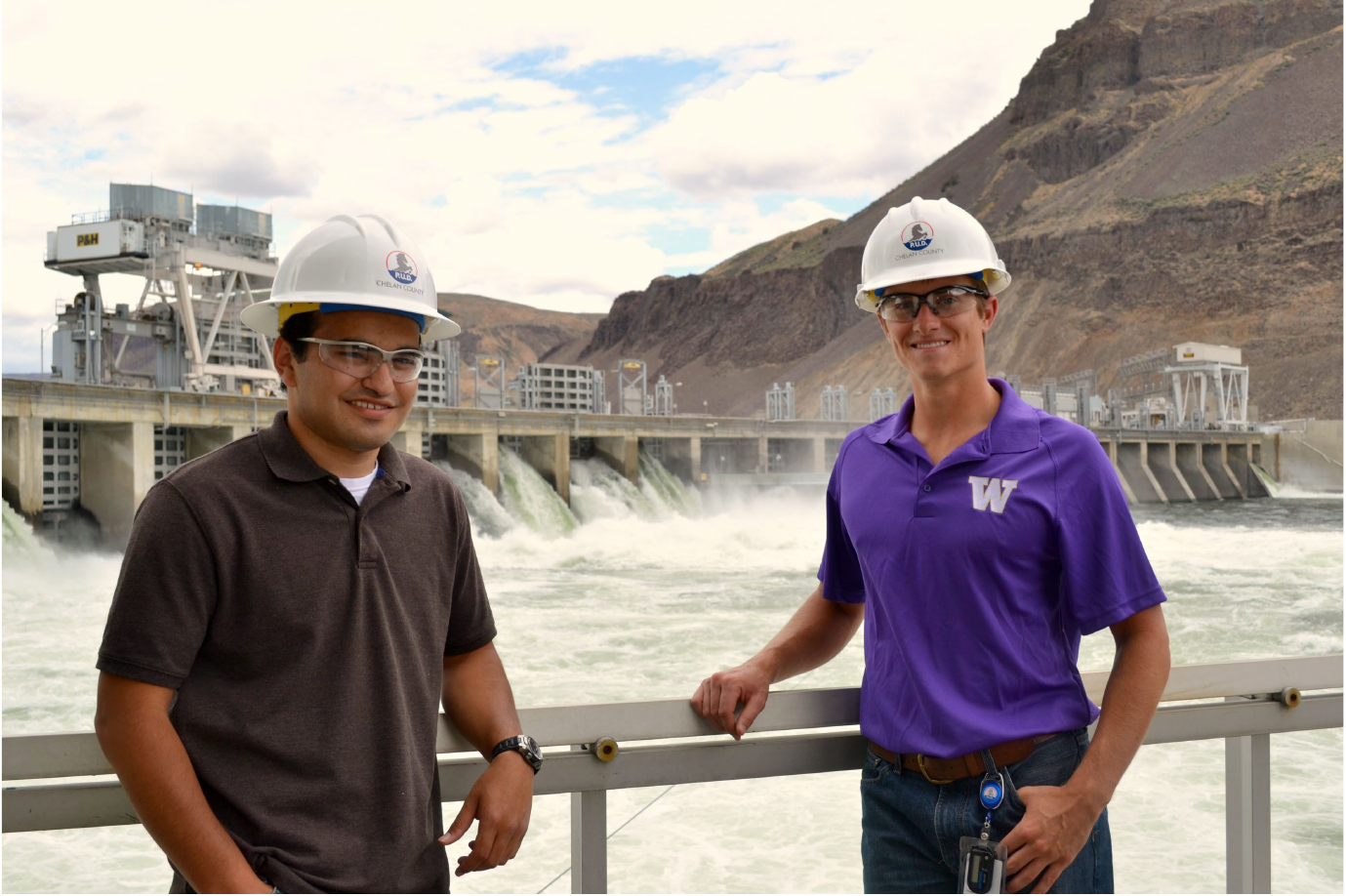}
	\label{fig:shwd_5}}
	\vfil
	\subfloat[]{\includegraphics[width=2.2in]{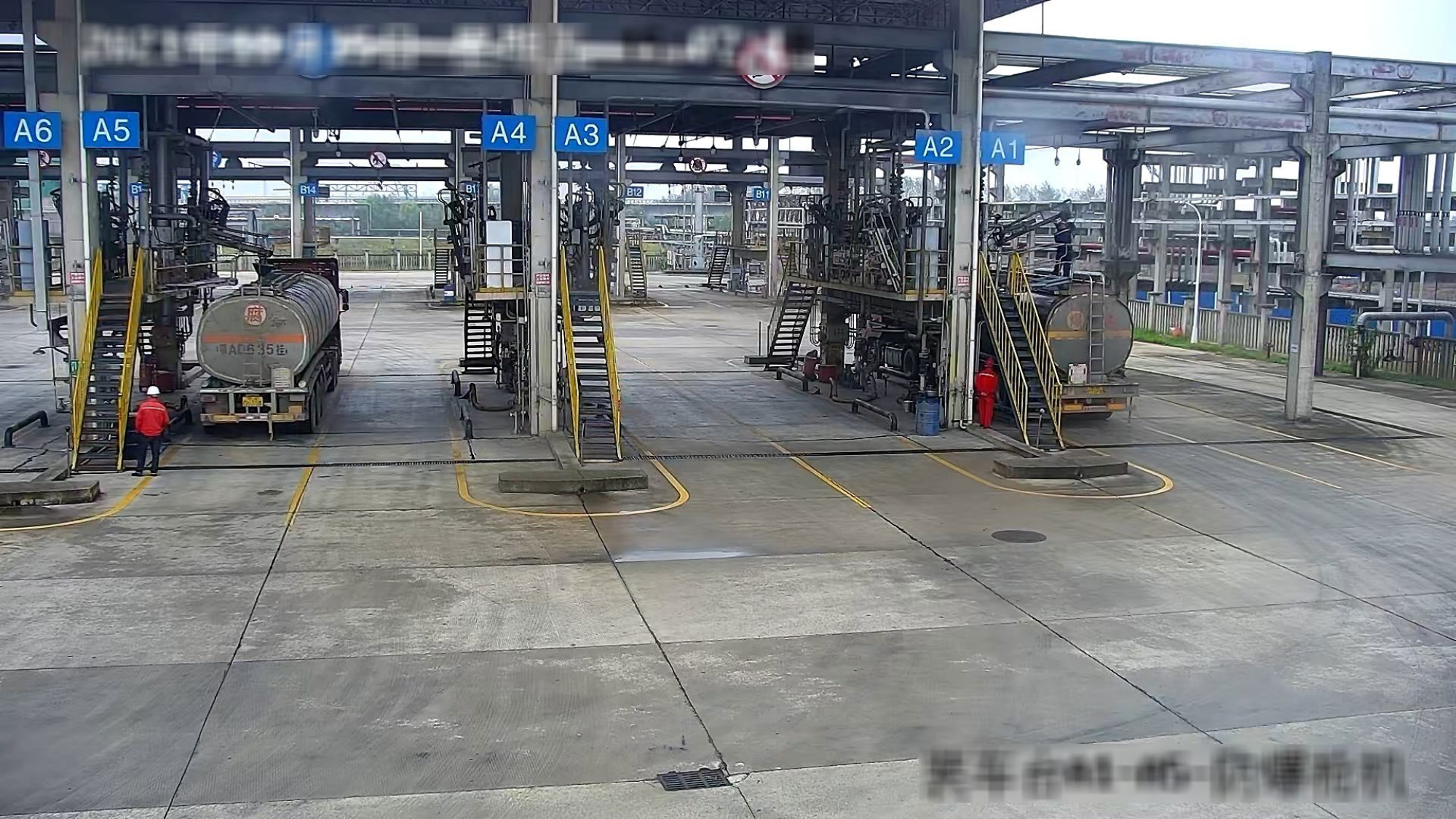}
	\label{fig:sfchd_1}}
	\hfil
	\subfloat[]{\includegraphics[width=2.2in]{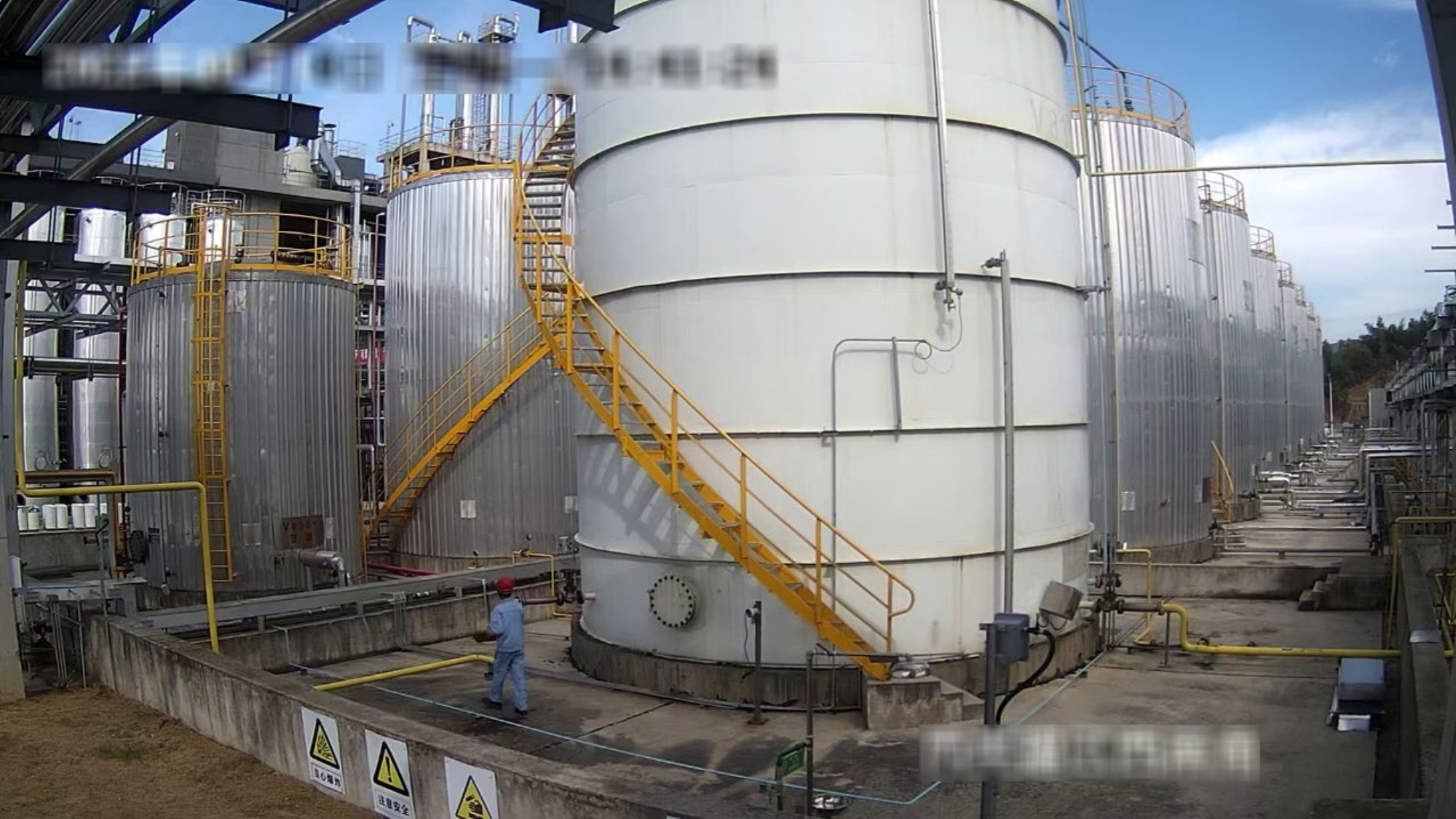}
	\label{fig:sfchd_2}}
	\hfil
	\subfloat[]{\includegraphics[width=2.2in]{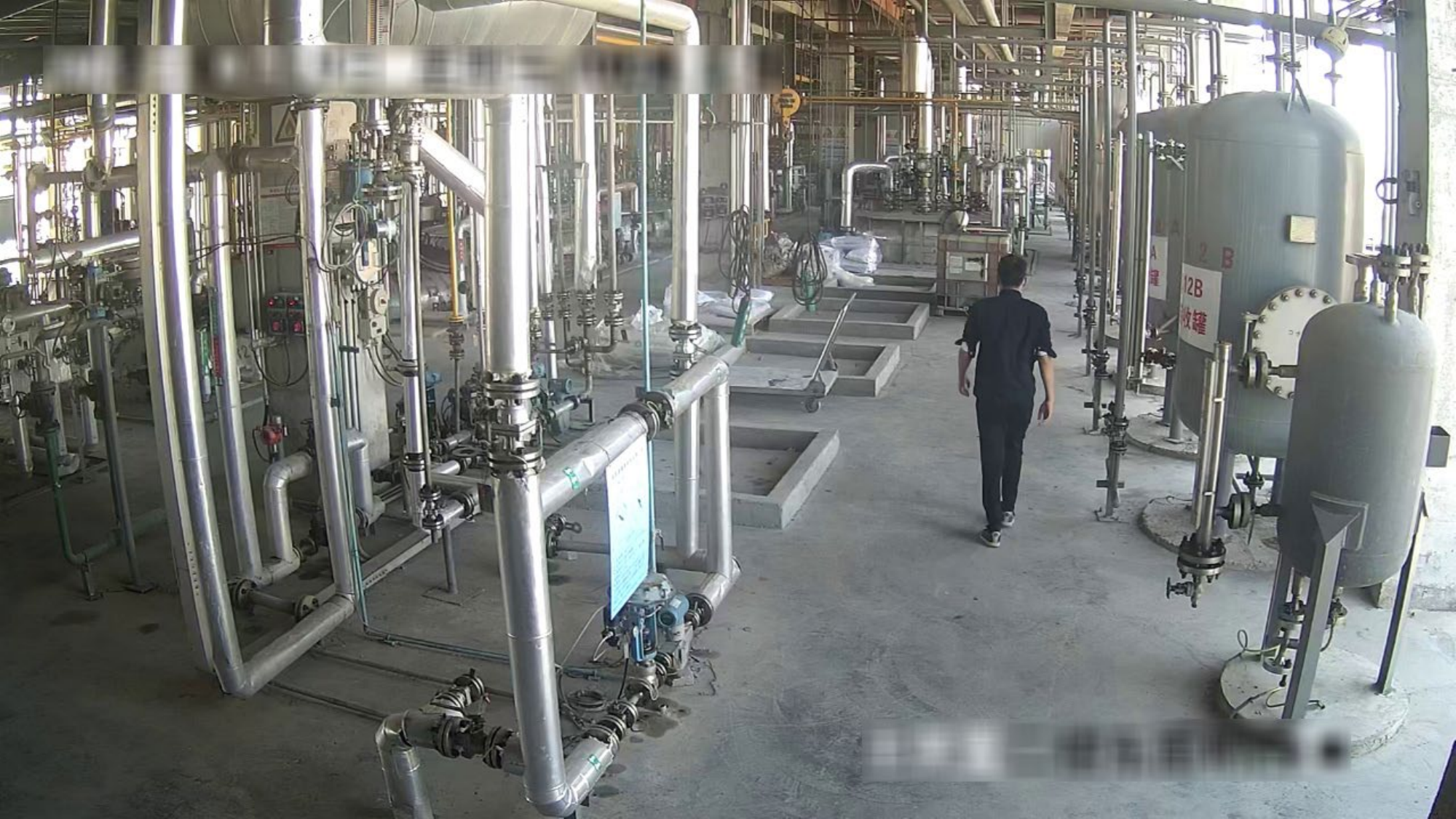}
	\label{fig:sfchd_3}}
	\caption{Examples of images from existing datasets and the SFCHD dataset. Here, (a)-(e), (f)-(j), and (k)-(m) are images selected from the Pictor-v3, SHWD, and SFCHD datasets, respectively.}
	\label{fig:example_dataset}
\end{figure*}

To address the above limitations, in this work, we contribute a large, complex, and realistic high-quality safety clothing and helmet detection (SFCHD) dataset. Fig.~\ref{fig:example_dataset}illustrates a comparison between the SFCHD dataset and existing datasets. It is evident that our SFCHD dataset offers more realistic work scenarios and more complex background environments. The SFCHD dataset comprises 12,373 images, covering 7 categories, with a total of 50,558 labeled instances, including labels such as Person, Safety Helmet, Safety Clothing, Other Clothing, and Head. To more closely mimic real-world work environments, we add two new labels for safety helmets and clothing under varying lighting conditions and shadow coverage: Blurred Clothing and Blurred Head. All images are captured from factory surveillance cameras, encompassing 40 different scenes across two chemical plants. Each instance has been manually annotated by professional inspectors, ensuring the accuracy of the annotations. It is worth noting that our SFCHD dataset not only provides a rich set of training samples but also serves as a benchmark for the evaluation of various detection tasks, such as small object detection, and high-low light object detection.

To more closely approximate real-world factory environments, our dataset includes a multitude of images captured under extreme lighting conditions, such as overexposure or underexposure, reflecting the actualities of the real world. When confronted with these extreme conditions, typical detection models often perform suboptimally. The reason is the presence of substantial noise in the images, which has a detrimental effect on the expression of potential features. To address this challenge and enhance the model's robustness in practical application scenarios, we introduce a spatial and channel attention-based low-light enhancement (SCALE) module. The SCALE module consists of two parallel pathways, i.e., the spatial attention pathway (SAP) and the channel attention pathway (CAP). SAP focuses on extracting features from key regions in the image, enhancing the model's perception of spatial feature relationships. CAP, on the other hand, assesses the importance of different channels, learning channel-specific feature information, assisting the model in extracting features such as color or texture from the image. SAP is composed of two scale-differentiated attention, achieving refined feature fusion and latent feature representation. The design of CAP is inspired by the channel attention mechanism in SENets~\cite{hu2018squeeze} and incorporates a skip connection structure. Unlike many image enhancement models that require specific loss functions, the proposed SCALE is a stand-alone, plug-and-play component, which can be seamlessly integrated into detection models and trained efficiently using only detection loss.

The main contributions of this work are as follows:
\begin{itemize}
	\item We contribute a large, complex, and realistic high-quality benchmark named SFCHD. To our knowledge, SFCHD is the inaugural factory dataset that concurrently satisfies the detection of both safety clothing and helmets.
	\item To address the challenge for image detection under low-light conditions, we propose a novel low-light enhancement module named SCALE. This plug-and-play module is composed of a spatial attention pathway and a channel attention pathway.
	\item We apply several classic object detection models on the SFCHD dataset for experimentation to verify its usability. Concurrently, we evaluate the SCALE module on both the ExDark and SFCHD datasets. The results substantiate that SCALE demonstrates significant enhancements in the detection of low-light images.
\end{itemize}

The rest of this paper is organized as follows. In Section~\ref{related_work}, we review the literatures pertinent to this work, encompassing the current state of datasets and the latest advancements of algorithms. In Section ~\ref{SFCHDdataset}, we provide a detailed account of the SFCHD dataset, including its construction principles, data details, and potential applications. We introduce the architecture of the SCALE module in Section~\ref{SCALEmodule}. In Section~\ref{experiments}, we initially apply several classical models to substantiate the SFCHD's usability, followed by an assessment of the SCALE's efficacy on the ExDark and SFCHD datasets. Finally, in Section~\ref{conclusion}, we summarize this work.

\section{Related Work}\label{related_work}
\subsection{Object Detection Datasets}
The most remarkable datasets in the field of object detection include Pascal VOC~\cite{everingham2010,everingham2015}, MS-COCO~\cite{lin2014microsoft}, ILSVRC~\cite{russakovsky2015}, and Open Image~\cite{kuznetsova2020}. The Pascal Vision Object Class (VOC) Challenge is a competition to accelerate the advancement of the field of computer vision, and two versions of the datasets adopted in the competition, VOC 2007 and VOC 2012, were commonly used for standard benchmarks. The Microsoft Common Objects in Context (MS-COCO) is one of the most challenging object detection datasets available, including two commonly used versions, COCO 2005 and COCO 2018. MS-COCO contains 80 categories, whose annotated objects are divided into three categories based on scale: small objects, medium objects, and large objects. The ImageNet Large-Scale Visual Recognition Challenge (ILSVRC) is also a competition that includes an object detection task, and its dataset serves as a benchmark for the evaluation of algorithm performance. The number of objects in the LISVRC was expanded to 200 compared to the VOC dataset, and the number of images and annotated objects are significantly larger than those in the VOC dataset. The Open Image dataset was launched by Google in 2017 and consists of 9.2 million images with 16 million bounding boxes for 600 categories on 1.9 million images used for detection, becoming one of the largest object detection datasets.

Despite the multitude of datasets~\cite{everingham2010,everingham2015,lin2014microsoft,russakovsky2015,kuznetsova2020} available for the field of object detection, there is a relative dearth of datasets specifically tailored to industrial scenarios. The availability of high-quality and authentic datasets for safety clothing and helmet detection is crucial for ensuring engineering safety and factory compliance. Although object detection algorithms have made significant strides, their widespread application in industrial settings is still limited due to the absence of high-quality datasets. To our knowledge, there are currently only two open-source datasets available for helmet detection, namely Pictor-v3~\cite{nath8} and SHWD~\cite{gochoo2021safety}. However, the utility of these datasets is constrained, and they struggle to meet the application demands of both academia and industry. The Pictor-v3 dataset contains only 1,330 annotated images, which is insufficient to meet the contemporary requirement for large-scale data in deep learning. Moreover, the dataset is deficient in the inclusion of negative samples, leading to an imbalanced data composition and a disproportionate emphasis on color labels. In reality, safety helmets come in a variety of colors, and an overemphasis on color labels could potentially impair the model's ability to recognize the physical attributes of the helmets. The SHWD dataset comprises 3,241 images related to safety helmets, with 10,457 instances annotated. Although there is a marked increase in the number of images compared to Pictor-v3, most images in SHWD are downloaded from the web, featuring relatively simple backgrounds that do not adequately represent the complexity of real-world scenarios. Furthermore, due to the diverse origins of the data, there is a significant variation in image resolution within the SHWD dataset, which may adversely affect the model's performance in practical applications. In this work, we collect a high-quality and large-scale dataset, SFCHD, that authentically reflects the complexity of industrial scenarios, to facilitate the application of object detection algorithms in the field of industrial safety. Table~\ref{tab:comparison} shows a comparison between our SFCHD and the extant safety helmet datasets.
\begin{table*}[htbp]
	\centering
	\setlength{\tabcolsep}{9pt}
	\caption{Comparison between SFCHD and existing open-source datasets for safety helmets}
	\label{tab:comparison}
	\begin{threeparttable}
	\begin{tabular}{ccccccccc}
	\toprule
	Dataset& Year &\#Category &\#Sample  &\#Instance&Color &Task &Data Source  \\
	\midrule
	Pictor-v3 & 2020 & 6 & 1330 & 9208 &RGB&Detection &Web-mined and Crowd-sourced \\
	SHWD & 2019 & 2 & 3241  & 10457 & RGB & Detection &Web-mined  \\
	\midrule
	SFCHD& 2023 &7 &12,373  &50558 &RGB &Detection &Chemical Plant  \\
	\bottomrule
	\end{tabular}
	In the SHWD dataset, only the images related to safety helmets are used. \#Category, \#Sample, and \#Instance denote the numbers of categories, samples, and instances in the dataset, respectively.
	\end{threeparttable}
\end{table*}

\subsection{Object Detection Methods}
In the field of computer vision, object detection is a fundamental task for many real-world applications, which are receiving widespread attention. There are many potential application scenarios for object detection, such as autonomous driving~\cite{feng2022review,9933424}, medical diagnosis~\cite{elakkiya2022cervical}, emotion detection~\cite{li2023ga2mif}, pedestrian detection~\cite{kim2021uncertainty,9612580}, etc. The early object detection algorithms are composed of hand-crafted feature extractors such as Histogram of Oriented Gradients (HOG)~\cite{dalal2005histograms}, Viola Jones detector~\cite{viola2001rapid}, and so on. These methods have inaccurate detection results and are resistant to be applied to different datasets. In recent years, as the rapid evolution of deep learning, the reintroduction of CNN based models has brought a new landscape to the realm of object detection. CNN-based~\cite{krizhevsky19,Xie23} object detection methods can be categorized into two main types: two-stage methods and one-stage methods. Two-stage algorithms have two separate steps, i.e., searching for object proposals in images during the first stage, and then classifying and localizing them during the second stage. These methods commonly are of complex architecture and demand longer time to generate proposals. Unlike the two-stage algorithms, one-stage approaches are superior in terms of real-time performance and network architecture. One-stage algorithms localize objects with the predefined boxes of different scales and aspect ratios, and classify semantic objects in a single shot adopting dense sampling~\cite{sahil2022survey}.

\begin{figure*}[htbp]
	\centering
	\includegraphics[width=6.8in]{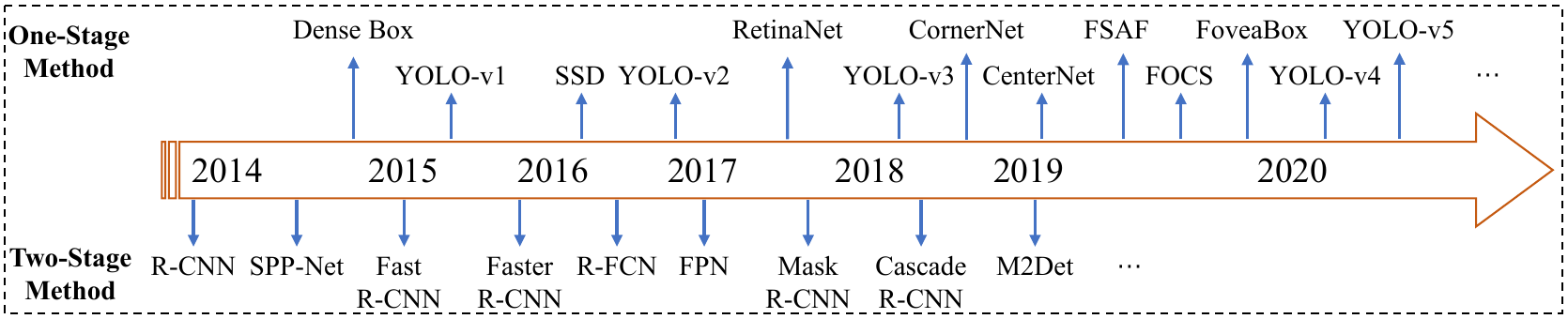}
	\caption{The developmental trajectory of object detection methods based on deep learning. The upper part illustrates the one-stage methods and the lower part showcases the two-stage methods.}
	\label{fig:development}
\end{figure*}

RCNN~\cite{girshick28} is a pioneering work in two-stage object detection. It employed a selective search algorithm to generate 2000 region proposals, then used a CNN to extract features from each proposal, followed by support vector machine (SVM)~\cite{platt27} and finally performed classification and regression. However, the slow inference speed due to the large number of region proposals limits the practicality of the model. Fast RCNN~\cite{Girshick25} further improved the speed and accuracy of RCNN~\cite{girshick28} by sharing features among proposals and using ROI-pooling to extract fixed-size features. Faster RCNN~\cite{ren26} unified the region proposal network and detection network, leading to improved accuracy and inference speed. One-stage methods, such as YOLO~\cite{redmon_v1_29}, directly predicted object categories and bounding boxes in a single network pass. YOLOv2~\cite{redmon_v2_30} improved the accuracy and speed by introducing batch normalization and full convolutional operations. YOLOv3~\cite{redmon_v3_31} further improved the accuracy by using a more powerful backbone network and multiscale prediction based on Darknet-53. YOLOv4~\cite{bochkovskiy_v4_32} introduced additional improvements in data preprocessing, network design, and prediction. SSD~\cite{liussd_33}, another one-stage method, detected objects of different scales by using feature maps from different layers. FCOS~\cite{tianfcos_34} predicted object boundaries and categories based on internal points and allocated objects to different pyramid levels based on their scales. YOLOv5 introduced some novel improvement ideas based on YOLOv4, which resulted in a significant performance advancement in both speed and accuracy. The development line of object detection algorithms based on deep learning is shown in Fig.~\ref{fig:development}.

\subsection{Low-Light Object Detection}
In low-light environments, the visibility of the image is severely compromised due to poor lighting conditions, greatly impacting the performance of detector. Initial methods for low-light detection focus on enhancing the low-light images before feeding them into detector. For instance, Zhang et al.~\cite{zhang2019Kind} introduced KinD, which achieves image enhancement through paired training with images at different levels of illumination. Lv et al.~\cite{Lv2018MBLLENLI} proposed a multi-branch low-light enhancement network (MBLLEN), which extracts features at various levels and then merges them using a multi-branch structure to produce enhanced images. Liu et al.~\cite{IA_yolo}proposed IA-YOLO, which uses a CNN-based parameter predictor to train the YOLO detector. Qin et al.~\cite{Qin_2022_ACCV_DENet} introduced DE-YOLO, by training the cascaded DENet and YOLOv3 models in an end-to-end manner with the standard detection loss. Cui~\cite{Cui_2022_BMVC_IAT} presented IAT-YOLO, which utilizes both local and global branches, incorporating transformers for image enhancement. Additionally, joint training methods have been employed, where only the loss from detection tasks is used to accomplish low-light detection.

\section{Proposed SFCHD Dataset}\label{SFCHDdataset}
Existing helmet datasets are insufficient to meet the requirements of practical scenarios. In this work, we collect real-world images from chemical plants and construct a high-quality safety clothing and helmets dataset for object detection. In the following, we introduce the construction principles, data details, and potential tasks of the proposed SFCHD dataset.

\subsection{Construction Principle}
To guarantee the high quality of the SFCHD dataset, we construct the SFCHD dataset in accordance with the following principles.

\textit{Realistic Source:}
Collecting realistic data is crucial for practical implementation and also makes the related studies more meaningful. Therefore, we collect on-site images of workers from surveillance cameras in two chemical plants based on specific collection criteria. For each camera, we save one image every 10 frames when there were people in the scene. This data collection process is carried out continuously for one week, resulting in approximately 40,000 captured images as the initial data source. Subsequently, we perform data filtering while ensuring an even distribution of scenes and categories. The goal is to avoid significant imbalances in the quantity of different categories. Eventually, we retain 12,373 images as the final dataset. We employ a repetitive filtering approach to ensure the authenticity and uniformity of the data.

\textit{Data Privacy:}
While collecting and labeling images, we strictly follow the standard de-identification procedure by removing any sensitive information (such as time, plant location, etc.), thereby ensuring that the identities of the factory and workers remain confidential.

\textit{Diverse Scenes:}
To ensure a diverse range of scenarios in the dataset, we collect data from two factories, capturing nearly 40 different indoor and outdoor scenes are captured. This approach ensures the dataset's scene universality.

\textit{Professional Annotation:}
Our annotation team consists of six professional security inspectors and the three graduate students who have undergone specialized training. Each instance in SFCHD is annotated by the six professional security inspectors. After annotation, three graduate students who are not involved in the labeling process were invited to review the annotations, ensuring that every image and every annotation in the dataset is completely accurate.

\textit{Various Categories:}
Our SFCHD dataset comprises five common categories observed at construction sites, including Person, Safety Helmet, Safety Clothing, Other Clothing, Head. To ensure the realism of the dataset, we add annotations for objects that are difficult to differentiate. In other words, we annotate Blurred Clothing and Blurred Head for those who are affected by shadows and strong lighting. In this manner, the SFCHD dataset encompasses a total of 7 categories, as depicted in Table~\ref{tab:distribution}.
\begin{table*}[htbp]
	\centering
	\setlength{\tabcolsep}{5pt}
	\caption{Statistics of instance distribution per category in the SFCHD dataset}
	\label{tab:distribution}
	\begin{tabular}{ccccccccc}
	\toprule
	Category & Person & Safety Helmet &  Safety Clothing & Other Clothing & Head & Blurred Clothing & Blurred Head & Total\\
	\midrule
	Training & 13,528 & 11,378  & 11,781 & 626 & 961 & 1,053 & 896 & 40,223\\
	Testing & 3,482 & 2,920  & 3,032 & 154 & 239 & 271 & 238 & 10,336 \\
	\midrule
	Total & 17,010 & 14,298  & 14,813 & 780 & 1200 & 1,324 & 1,134 & 50,559\\
	\bottomrule
	\end{tabular}
\end{table*}

\begin{figure*}[htbp]
	\centering
	\subfloat[]{\includegraphics[width=2.2in]{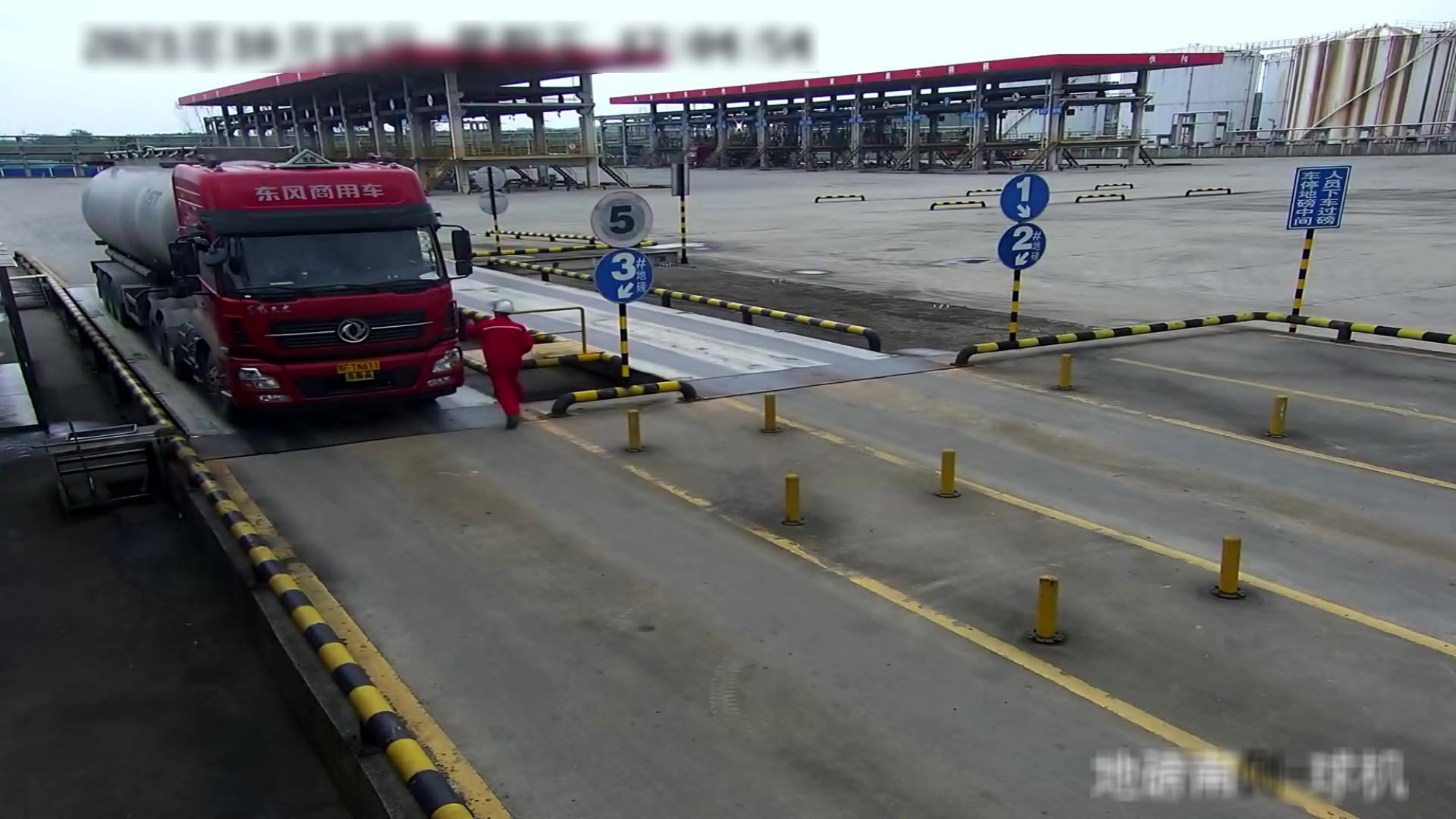}%
	\label{fig:normal_1}}
	\hfil
	\subfloat[]{\includegraphics[width=2.2in]{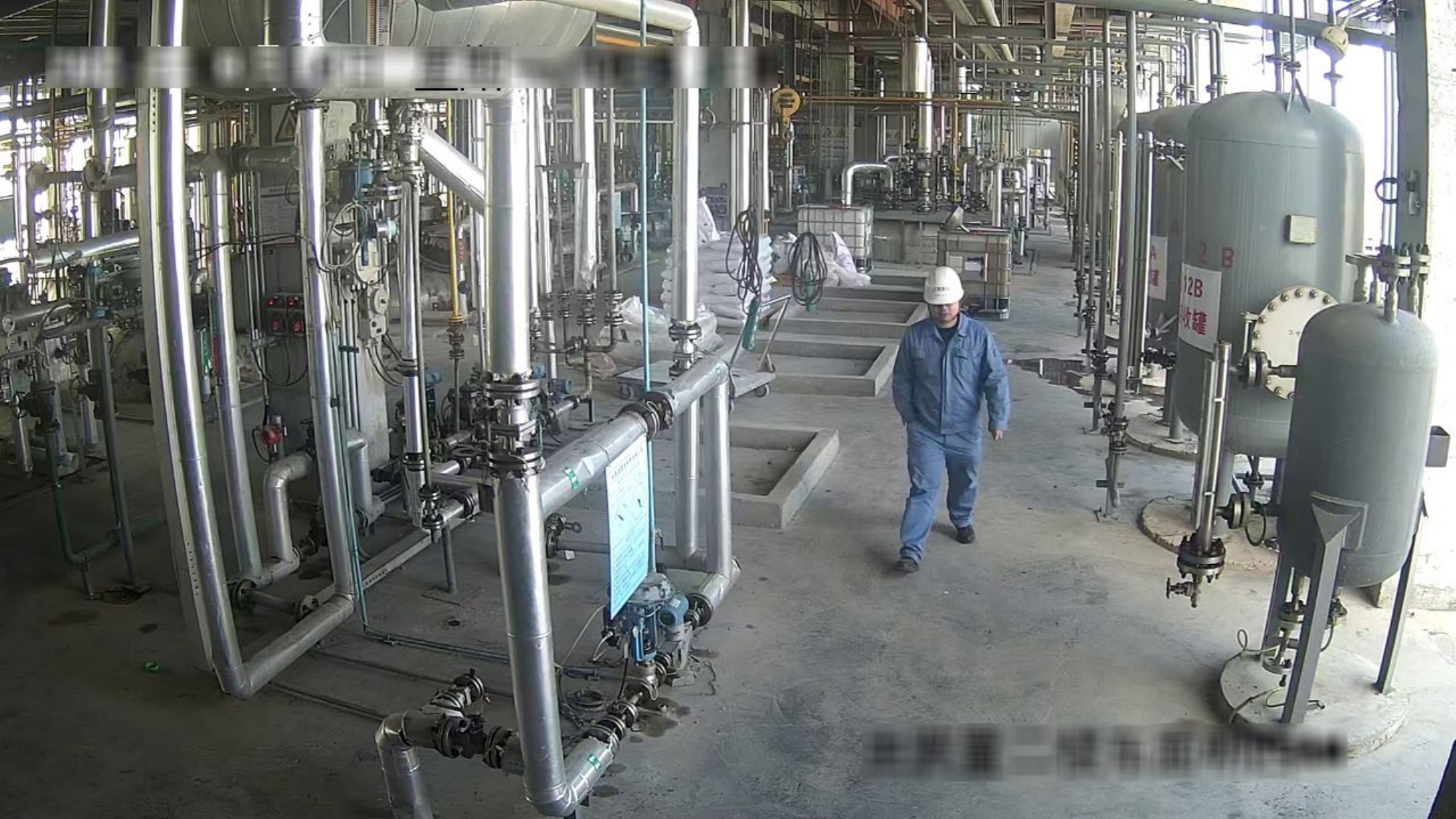}%
	\label{fig:normal_2}}
	\hfil
	\subfloat[]{\includegraphics[width=2.2in]{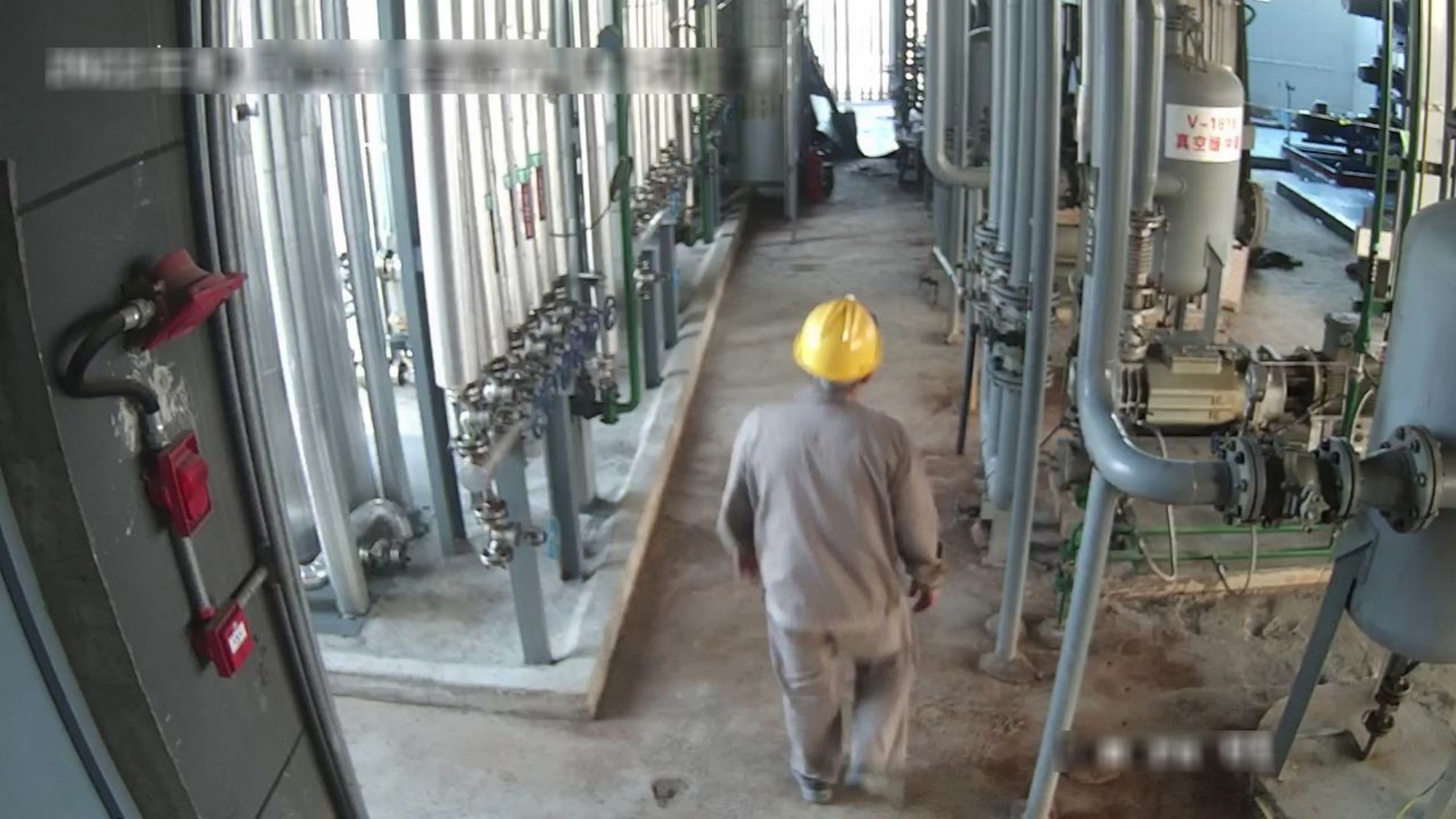}%
	\label{fig:normal_3}}
	\vfil
	\subfloat[]{\includegraphics[width=2.2in]{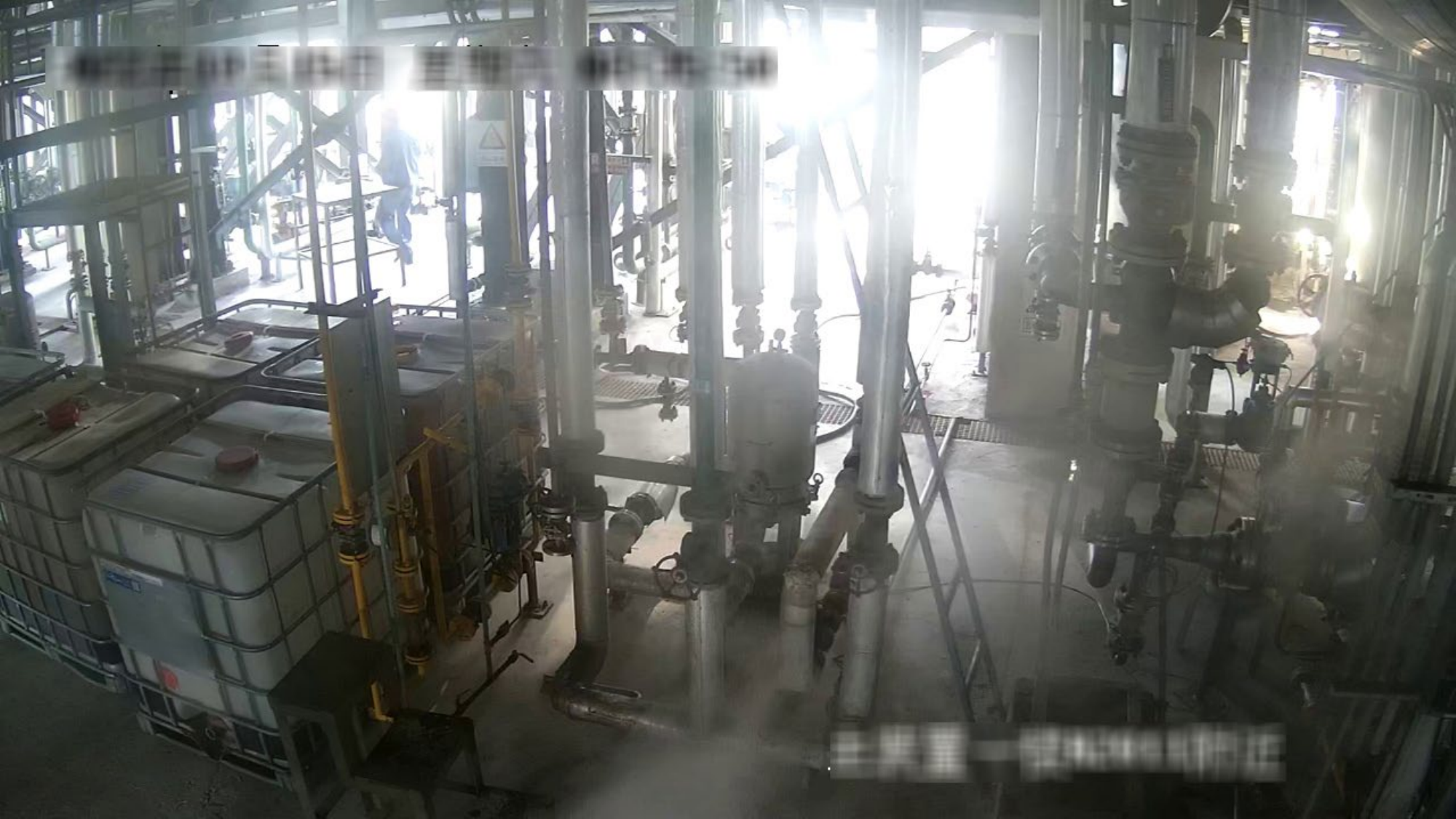}%
	\label{fig:high_1}}
	\hfil
	\subfloat[]{\includegraphics[width=2.2in]{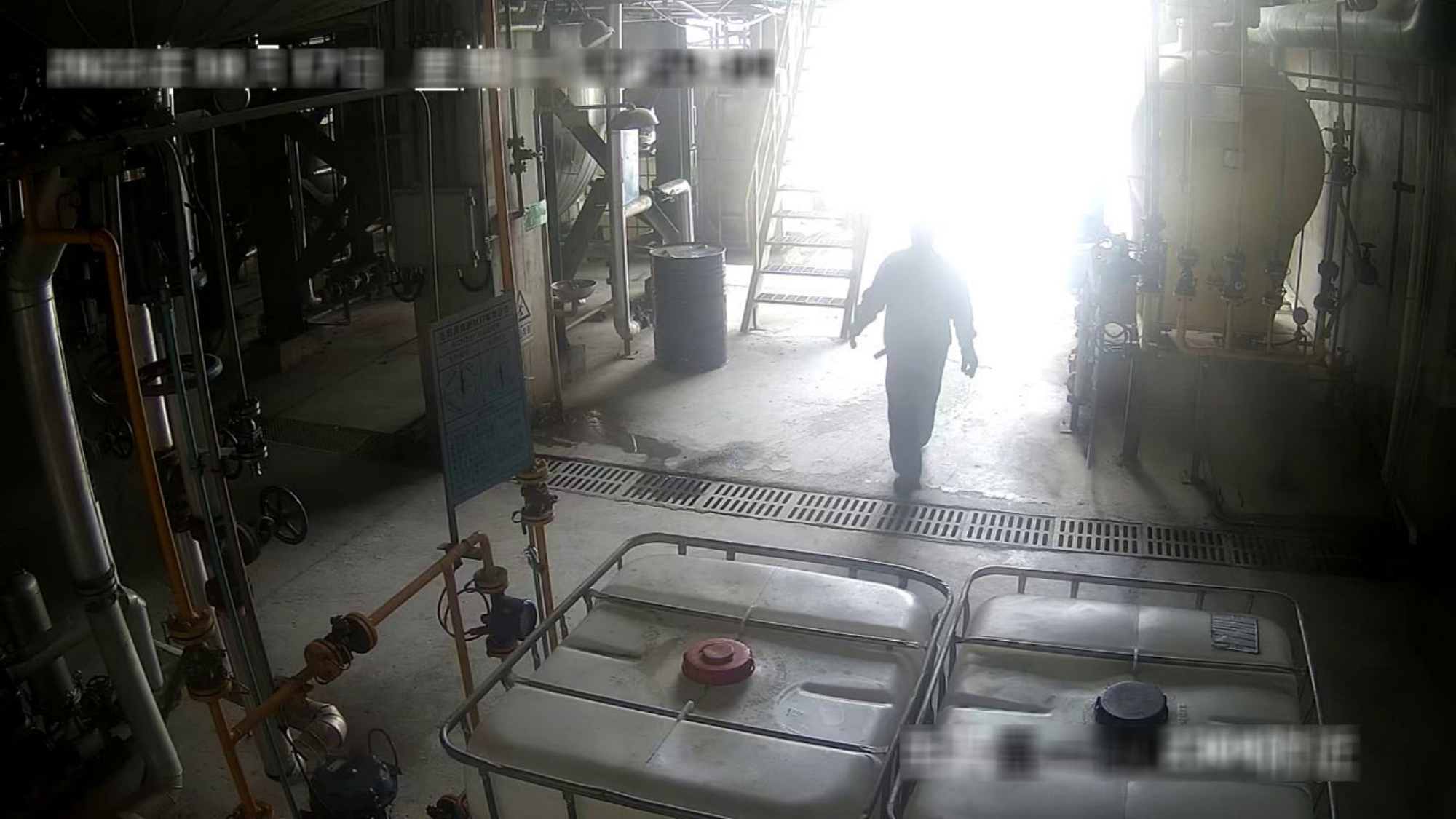}%
	\label{fig:high_2}}
	\hfil
	\subfloat[]{\includegraphics[width=2.2in]{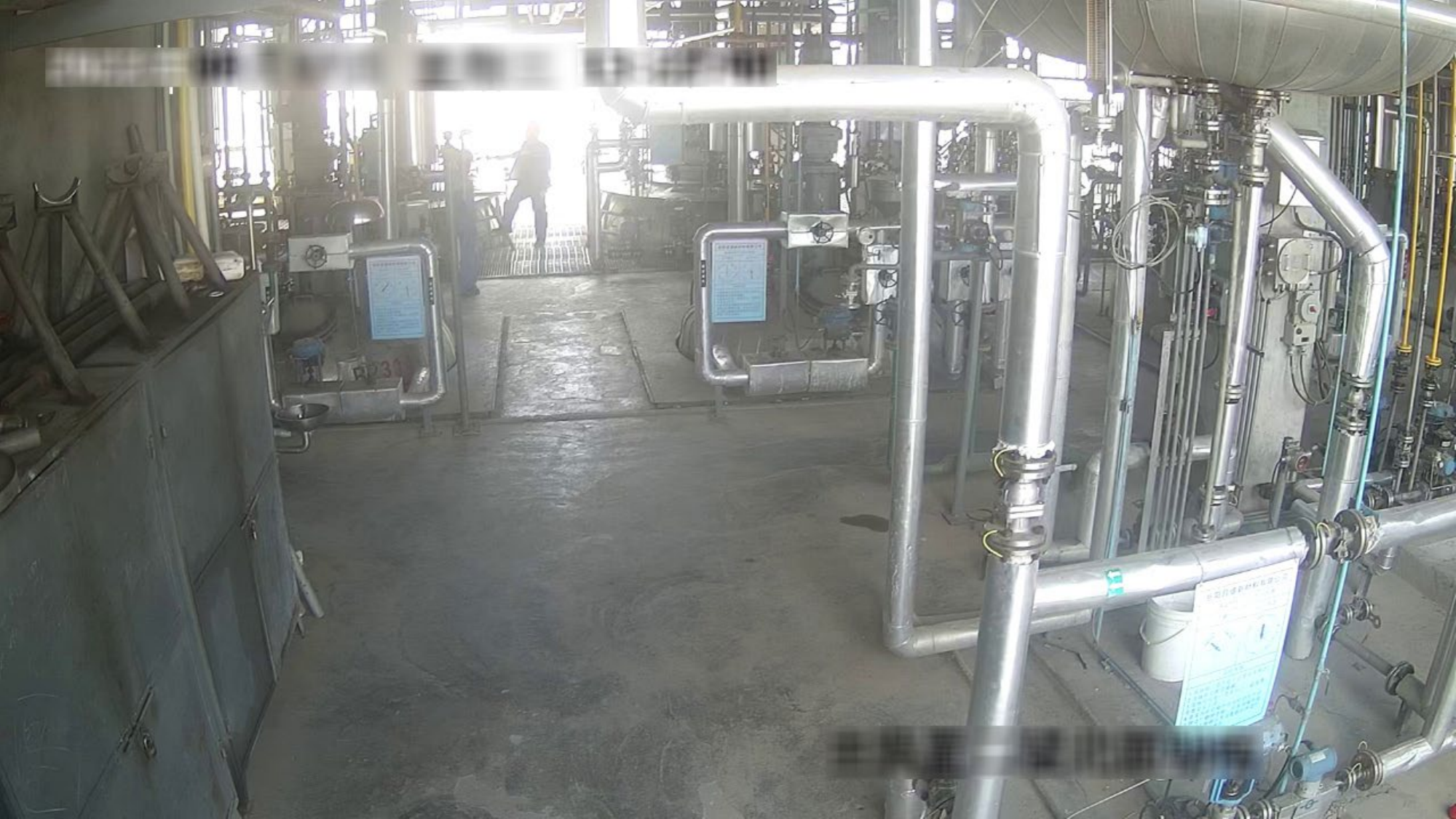}%
	\label{fig:high_3}}
	\vfil
	\subfloat[]{\includegraphics[width=2.2in]{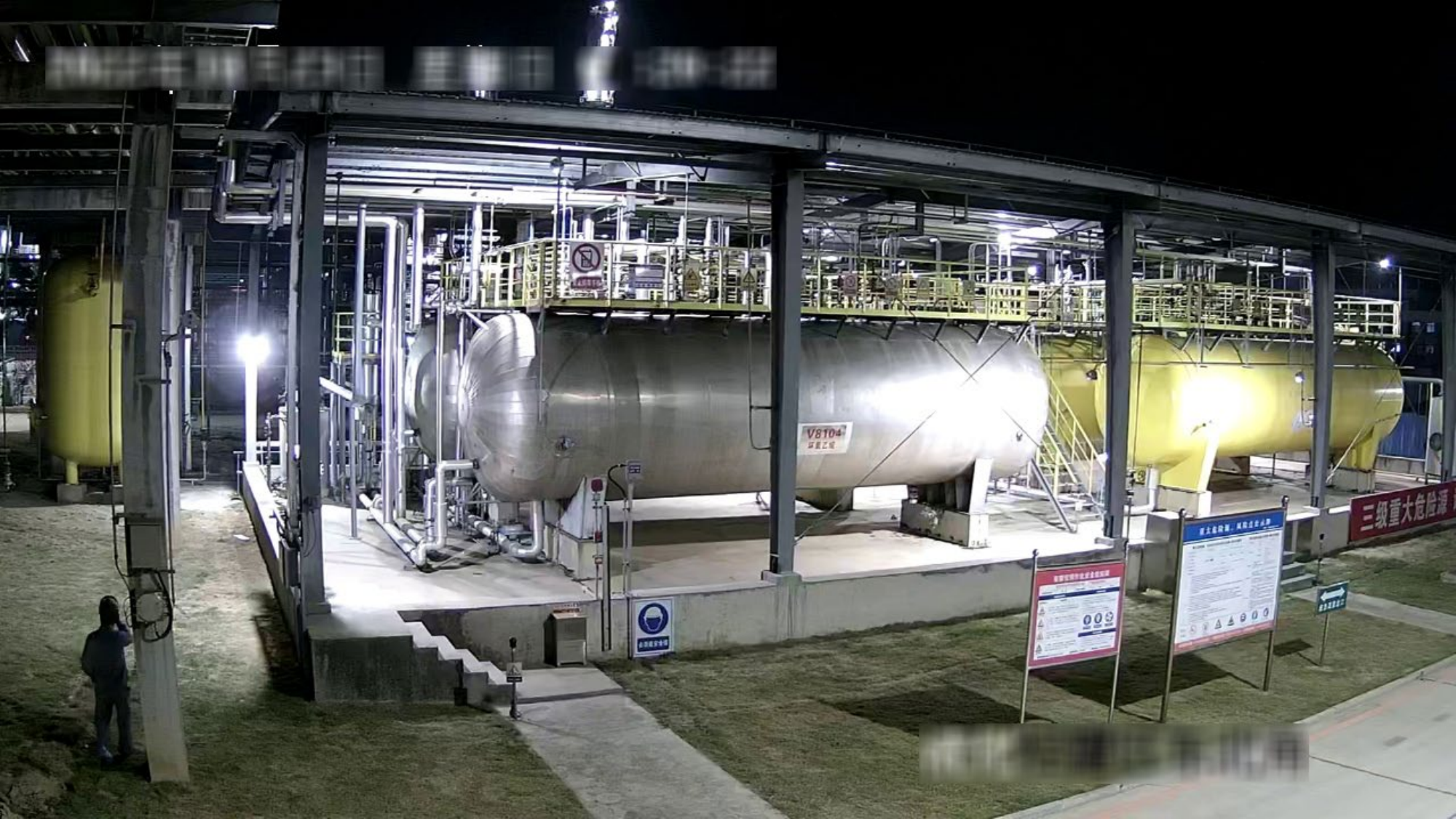}%
	\label{fig:low_1}}
	\hfil
	\subfloat[]{\includegraphics[width=2.2in]{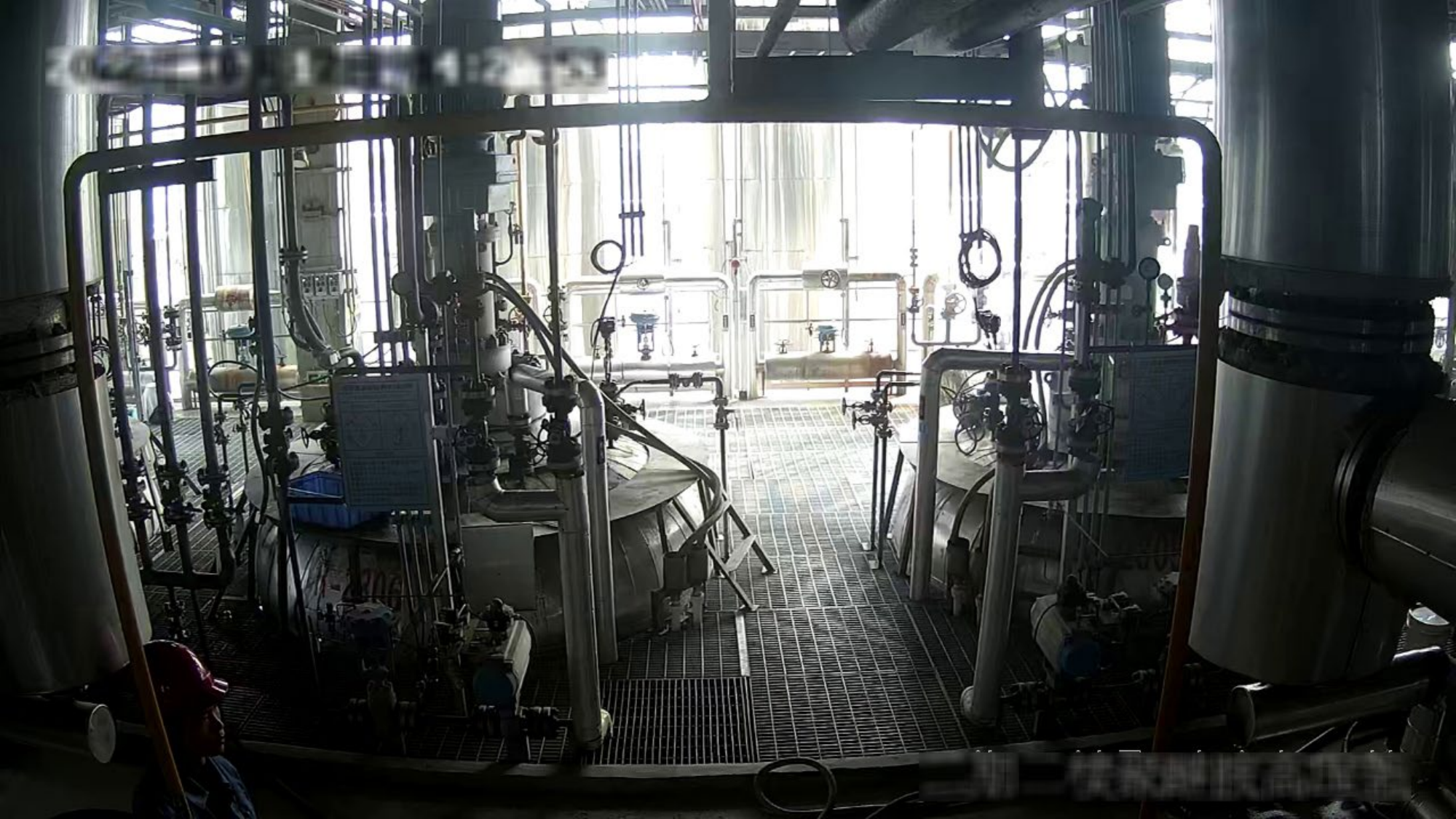}%
	\label{fig:low_2}}
	\hfil
	\subfloat[]{\includegraphics[width=2.2in]{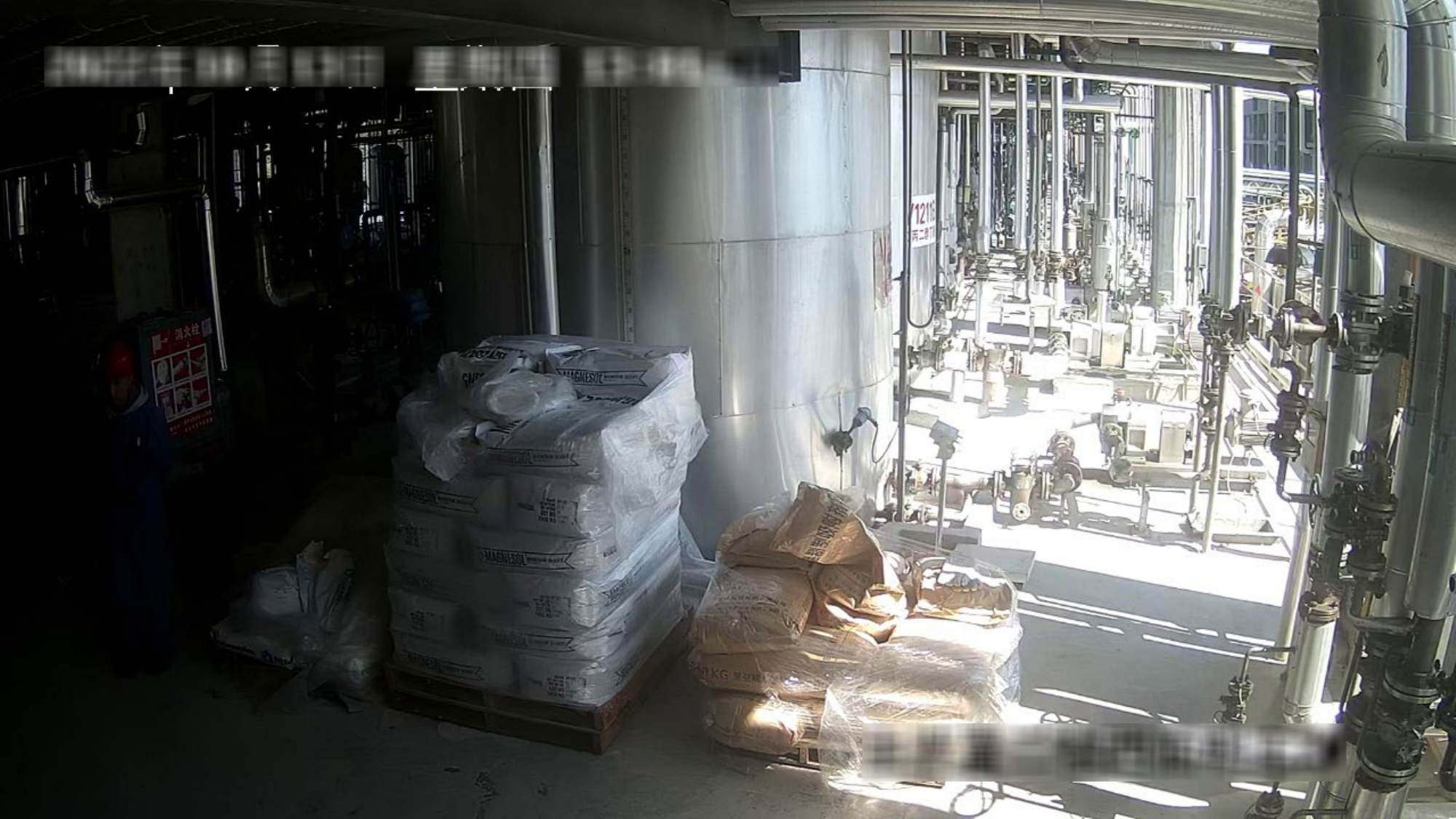}%
	\label{fig:low_3}}
	\caption{Examples of images with different lighting conditions. Here, (a)-(c), (d)-(f), and (g)-(i) are the images based on normal-light, high-light, and low-light, respectively. In (d)-(f), it is extremely difficult to detect whether workers are wearing safety clothing and helmets because the surrounding light is too high. Similarly, in (g)-(i), too low light around workers makes it difficult to conduct object detection tasks.}
	\label{fig:light_image}
\end{figure*}

\textit{High-Low Light Scene Discriminate:}
As illustrated in Fig.~\ref{fig:light_image}, we categorize the entire dataset into three groups: high-light, low-light, and normal-light images, based on the intensity of light sources around person in the images. Images with an evenly distributed light source are classified as normal-light. On the other hand, images with apparent strong light sources or shadows are classified as high-light and low-light, respectively. Although deep learning-based object detection methods have exhibited impressive results on conventional datasets, detecting objects accurately in low-quality images captured under extreme light conditions remains a significant challenge. Thus, we provide researchers with samples of extreme light conditions to encourage the development of object detection in such scenarios. The classification of all scenes is carried out by the three graduate students. Given the potential for different perspectives on classifying images, a voting system was used to resolve any disagreements, with the majority opinion determining the final classification outcome.

\subsection{Dataset Details}
In accordance with the construction principles of the dataset, we amass a large, complex, and realistic dataset for safety clothing and helmets, known as SFCHD. Subsequently, we will provide a description of the SFCHD dataset in terms of instance distribution, data partitioning, and data quality.

\begin{figure}[htbp]
	\centering
	\includegraphics[width=3.0in]{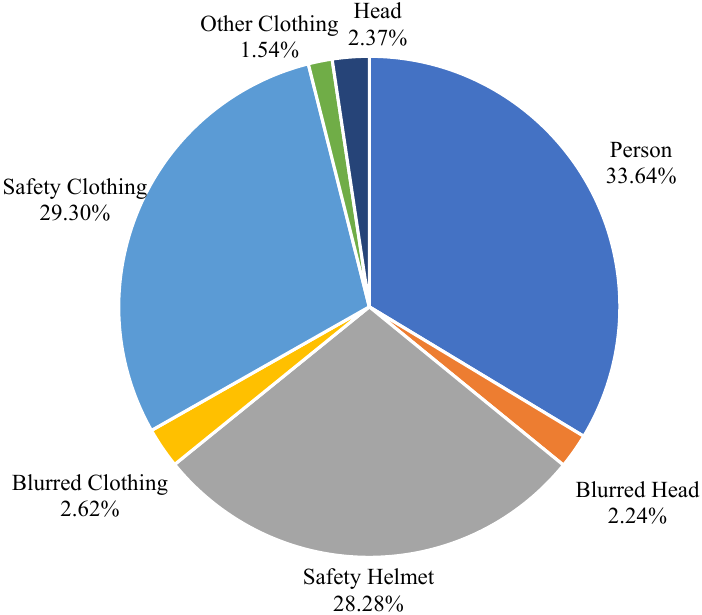}
	\caption{The proportion of instances per category in the SFCHD dataset.}
	\label{fig:proportion}
\end{figure}

\textit{Instance Distribution:}
The SFCHD dataset comprises 12,373 images, 7 categories, and 50,552 annotations. The instance proportion of each category is shown in Fig.~\ref{fig:proportion}. As depicted in Fig.~\ref{fig:proportion}, the most prominent categories are Person, Safety Helmet, and Safety Clothing, accounting for about 33.64\%, 28.28\% and 29.30\% of the dataset, respectively. Due to the rarity of violations in real-world chemical plants, situations where safety clothing or helmets are not worn are infrequent. As a result, the instances classified as Other Clothing and Head represent only a small portion of the dataset, 1.54\% and 2.37\%, respectively. The remaining categories, Blurred Clothing and Blurred Head, which represent instances captured in special environments, are relatively limited in quantity, accounting for 2.24\% and 2.62\%, respectively.

\textit{Data Partitioning:}
Our dataset is divided into training set and testing set, and the ratio is 4:1. We have 12,373 images in SFCHD dataset, including 9,898 images in the training set, 2,475 images in the testing set. The statistics of category distribution of training and testing set are also shown in Table~\ref{tab:distribution}.

\textit{Data Quality:}
To facilitate model training, all images are stored in JPG format; the average resolution of the images is predominantly 1280 * 704, with the maximum resolution reaching 1920 * 1280.

\subsection{Potential Tasks}
In addition to conventional object detection, the SFCHD dataset can also serve as a benchmark for the assessment of various extreme detection tasks, such as small object detection and high-low light object detection.

\textit{Small Object Detection:}
As per the definition provided by the international organization SPIE, a small object must not exceed 0.12\% of the overall image area. Therefore, we define small objects as those whose actual bounding box occupies less than 0.1\% of the entire image area, while large objects are those whose bounding box occupies more than 0.2\% of the image area. The remaining objects are classified as medium-sized objects. Table~\ref{tab:small_objects} provides detailed statistics on the number of small objects in the dataset.
\begin{table}[htbp]
\centering
\setlength{\tabcolsep}{7pt}
\caption{Category distribution of objects with different sizes in the SFCHD dataset
}
\label{tab:small_objects}
% \begin{threeparttable}
\begin{tabular}{cccccccccc}
\toprule
Category  & Total & Large & Medium & Small \\
\midrule
Safety Helmet & 14,298 & 3,551  & 3,030 & 7,717 \\
Head & 1,200 & 95  & 164 &941 \\
Blurred Clothing & 1,324 & 281 & 313 & 730 \\
Blurred Head & 1,134 & 15  & 41 & 1,078 \\
\bottomrule
\end{tabular}
% \end{threeparttable}
\end{table}

\begin{figure}
	\centering
	\includegraphics[width=3.0in]{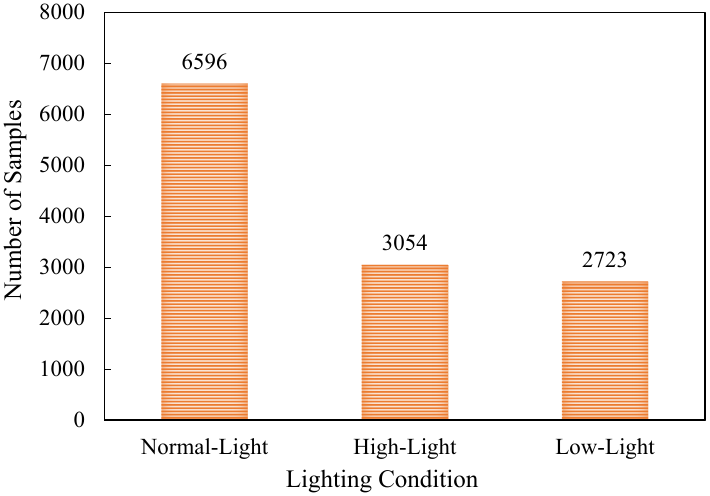}
	\caption{The quantity distribution under different lighting conditions in the SFCHD dataset.}
	\label{fig:light}
\end{figure}

\textit{High-Low Light Detection:} 
A critical reason why object detection tasks are challenging under extreme light conditions is the lack of relevant datasets. According to the light conditions of the area where the object is located, we divide the images in the dataset into high-light image, low-light image and normal-light image. Fig.~\ref{fig:light_image} illustrates the quality of each type of image. The quantity distribution is shown in Fig.~\ref{fig:light}, and we can conclude that the samples under normal-light are the most, while the samples under high- and low-light are basically the same.

\section{Proposed SCALE Module}\label{SCALEmodule}
\begin{figure*}[htbp]
	\centering
	\includegraphics[width=6.8in]{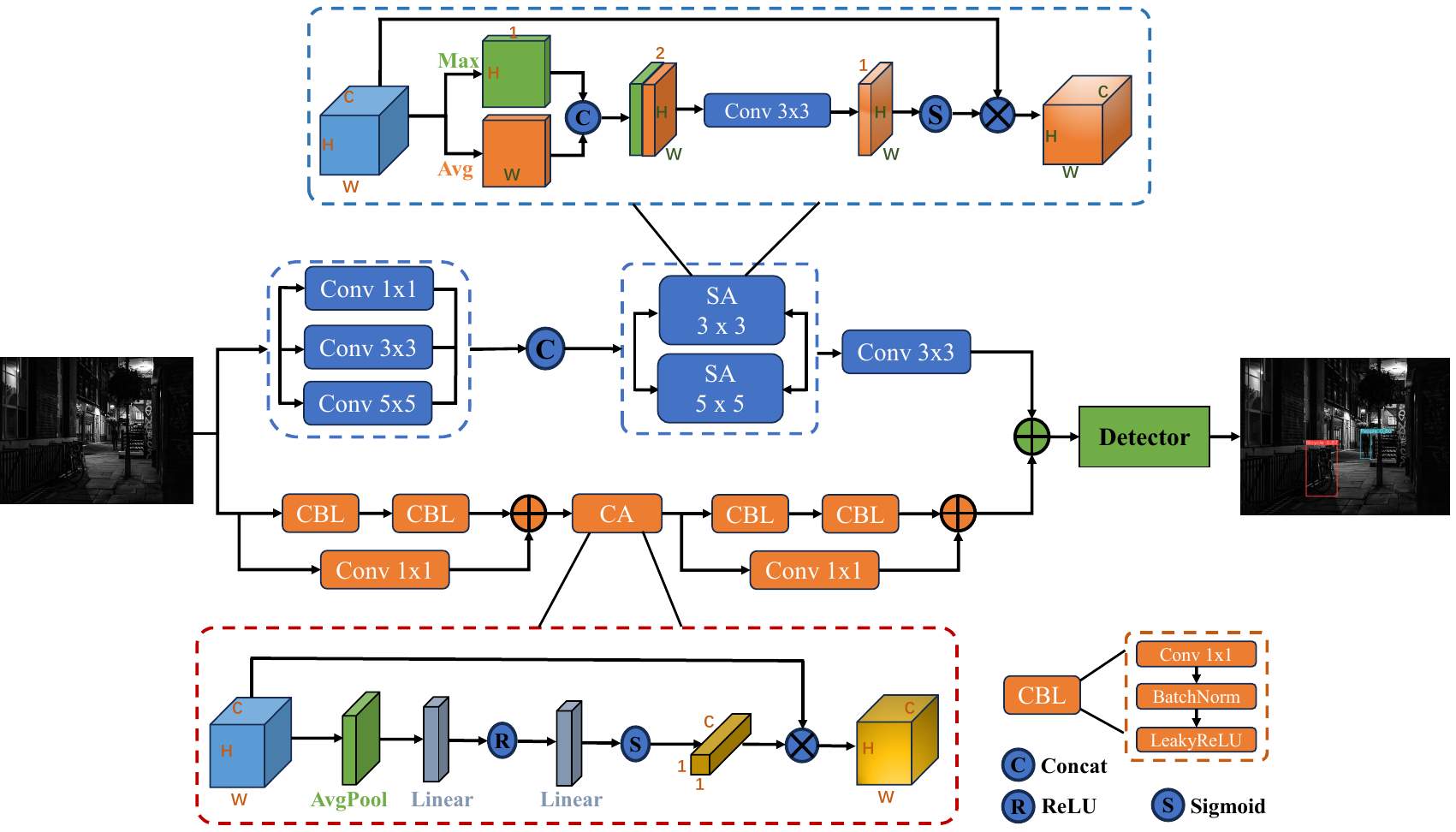}
	\caption{Network structure of spatial and channel attention-based low-light enhancement (SCALE) module.}
	\label{fig:module}
\end{figure*}

To address the challenges faced by object detection models in low-light environments, we design a spatial and channel attention-based low-light enhancement (SCALE) module. SCALE, as a plug-and-play component, integrates seamlessly into traditional object detection workflows, as depicted in Fig.~\ref{fig:module}. Specifically engineered for extreme low-light conditions, the SCALE module adaptively enhances images, consisting primarily of two components, i.e., spatial attention pathway (SAP) and channel attention pathway (CAP). SAP enhances the network's ability to focus on important areas of the image by assigning differentiated weights to pixels in various locations, thereby more effectively understanding and capturing the spatial structure and its correlations within the image. The CAP, on the other hand, guides the network to concentrate on the most informative and significant channels by assigning weights, while suppressing features from redundant channels, further improving the quality and accuracy of feature expression. Features processed by the SCALE module are then fed into the object detector. The entire pipeline relies solely on detection loss for end-to-end training, simplifying the training process and enhancing efficiency. This design not only improves the model's performance under low-light conditions but also maintains the simplicity of model training.

\subsection{Spatial Attention Pathway}
For the SAP branch, the low-light image $I$ is initially input into three convolutional kernels of varying sizes to extract features from multiple scales. Subsequently, the obtained feature maps are concatenated along the channel dimension to enrich the feature space. This process can be mathematically represented as:
\begin{equation}
\begin{aligned}
	&F_{k} = \mathrm{Conv}_{k}(I), \quad k \in \{1,3,5\}, \\
	&F_{Conv} = \mathrm{Cat}[F_{1}, F_{3}, F_{5}],
\end{aligned}
\end{equation}
where $I$ denotes an input with a resolution of $H \times W \times C$; $\mathrm{Conv}_k$ denotes the convolutional operation with a kernel of size $k$, e.g., $\mathrm{Conv}_1$ is the convolution with a kernel of size $1$; $F_{k}$ denotes the corresponding convolution branch output; $\mathrm{Cat}$ denotes the concatenation operation across the channel dimension; $F_{Conv}$ denotes the feature map after passing through parallel convolution branches.

The obtained feature map $F_{Conv}$ is then directed into two independent spatial attention units (i.e., SA in Fig.~\ref{fig:module}). Each Spatial Attention unit primarily consists of pooling, convolution, and element-wise multiplication operations. Initially, we apply global max pooling and global average pooling to the feature map. The global max pooling helps the network to filter the most salient feature, focusing on areas of the low-light image that contain key information while disregarding the effects of darkness and noise. The global average pooling smooths the feature value, making the extracted feature more stable and representative. Subsequently, these two pooled feature maps are concatenated, fed into a convolution for fusion, and normalized through a sigmoid function. Ultimately, the normalized fused feature is element-wise multiplied with each channel of the original feature map, yielding the weighted spatial attention feature maps $F_{SA_k}$ ($k \in \{3,5\}$). The process above can be mathematically expressed as:
\begin{equation}
	F_{SA_k} = F_{Conv} \otimes \sigma(\mathrm{Conv}_{k}(\mathrm{Cat}[F_{Max}, F_{Avg}])),
\end{equation}
where $F_{Max}$ and $F_{Avg}$ represent the values after applying max pooling and average pooling to $F_{Conv}$, respectively; $\sigma$ represents the sigmoid activation function, and $\otimes$ is the element-wise multiplication. The spatial attention feature maps $F_{SA_3}$ and $F_{SA_5}$ are concatenated, then passed through a convolution of size 3 to produce the output of the SAP branch:
\begin{equation}
\begin{aligned}
	&F_{Conv}^\prime = \mathrm{Conv}_{3}(\mathrm{Cat}[F_{SA_3}, F_{SA_5}]),\\
	&I_{SAP} =  I \odot F_{Conv}^\prime,
\end{aligned}
\end{equation}
where $\odot$ denotes Hadamard multiplication, $I_{SAP}$ represents the output of SAP branch.

\subsection{Channel Attention Pathway}
For CAP branch, before entering the channel attention unit (CA in Fig.~\ref{fig:module}), the low-light image $I$ is first input into a residual block. This residual block primarily consists of convolution, batch normalization, and residual operations, which can be mathematically expressed as:
\begin{equation}
\begin{aligned}
& \hat{F} _{Conv} = \mathrm{Conv}_1(I), \\
& F_{CBL} = \mathrm{LR}(\mathrm{BN}(\mathrm{Conv}_1(I))), \\
& F_{Res} = \hat{F} _{Conv} + \mathrm{LR}(\mathrm{BN}(\mathrm{Conv}_3(F_{CBL}))),
\end{aligned}
\end{equation}
where $\mathrm{Conv}_1$ and $\mathrm{Conv}_3$ denote convolution operations with kernel sizes of 1 and 3, respectively; $\mathrm{LR}$ represents the LeakyReLU activation function, and $\mathrm{BN}$ denotes batch normalization; a residual block comprises two CBL blocks, and each CBL block includes a convolution layer, LeakyReLU function, and batch normalization; $F_{CBL_1}$ signifies the feature map after the first CBL block, and $F_{Res}$ indicates the final output of the residual block.

Inspired by SENet~\cite{hu2018squeeze}, the obtained feature map $F_{Res}$ is then directed into the channel attention unit. This unit primarily consists of pooling, linear transformation, and element-wise multiplication operations. Initially, an average pooling operation is employed to down-sample the feature map. This operation not only reduces the complexity of the feature map but also diminishes noise and detail interference, capturing local statistical information. Subsequently, the pooled feature map undergoes two linear transformations and is normalized through a sigmoid function. Ultimately, the resulting feature map is element-wise multiplied with the original feature map across channels to yield the channel attention feature map $F_{CA}$. This process assigns different weights to each feature channel, enabling the network to learn the most informative feature channels. The computation can be mathematically represented as:
\begin{equation}
F_{CA} = F_{Res}  \otimes  \sigma(\mathrm{Lin}_2(\mathrm{ReLU}(\mathrm{Lin}_1(F_{Avg}^\prime)))),
\end{equation}
where $F_{Avg}^\prime$ denotes the value after applying the average pooling operation to $F_{Res}$; $\mathrm{Lin}_1$ and $\mathrm{Lin}_2$ represent two successive linear transformations. $F_{CA}$ is processed through a residual block to obtain the output of the CAP branch:
\begin{equation}
	\begin{aligned}
	& \tilde{F} _{Conv}  = \mathrm{Conv}_1(F_{CA}), \\
	& \tilde{F}_{CBL} = \mathrm{LR}(\mathrm{BN}(\mathrm{Conv}_1(F_{CA}))), \\
	& \tilde{F}_{Res} = \tilde{F}_{Conv} + \mathrm{LR}(\mathrm{BN}(\mathrm{Conv}_3(\tilde{F}_{CBL}))),\\
	& I_{CAP} = I \odot \tilde{F}_{Res},
\end{aligned}
\end{equation}
where $\odot$ denotes Hadamard multiplication, $I_{CAP}$ denotes the output of CAP branch.

\section{Experiment}\label{experiments}
Centered around the SFCHD dataset and the SCALE module, our experiments are composed of three parts. Initially, we utilize a variety of classic detection methods to perform experiments on the existing helmet datasets and our SFCHD dataset, validating the practicality of our dataset. Subsequently, we incorporate the SCALE module into detection algorithms and conduct experiments on the public low-light dataset to confirm the effectiveness of the SCALE module. Finally, we integrate the SCALE module with the SFCHD dataset, demonstrating its capability to significantly improve detection accuracy under challenging conditions.

\subsection{Usability of the SFCHD Dataset}
In this subsection, we employ several detection models to conduct experiments on the existing helmet datasets and proposed SFCHD dataset, reporting their outcomes to substantiate the usability of our dataset.

\subsubsection{Experimental Setup}
\textit{Datasets:} We compare the propose SFCHD with the Pictor-v3~\cite{nath8} and SHWD~\cite{gochoo2021safety} datasets. Pictor-v3 contains 774 crowd-sourced and 698 web-mined images, where the crowd-sourced and web-mined images contain 2,496 and 2,230 instances of workers, respectively. SHWD provides the dataset used for both safety helmet wearing and human head detection. It includes 7,581 images with 9,044 human safety helmets wearing objects (positive) and 111,514 normal head objects (not wearing or negative).

\textit{Methods:} We utilize classic detection methods to verify the usability of our dataset, including SSD~\cite{liussd_33}, Faster RCNN~\cite{ren26}, RetinaNet~\cite{linfocal38}, FCOS~\cite{tianfcos_34}, TOOD~\cite{fengtood37}, VFNet \cite{zhang36}, YOLOv5\footnote{https://github.com/ultralytics/yolov5}. SSD was a one-stage, multi-scale detection algorithm that utilized a convolutional pyramid for feature extraction and performed classification and regression across feature maps. Faster RCNN was a pioneering two-stage object detection framework that identified potential object regions, or anchor boxes, and then precisely classified and localized the objects within these regions. RetinaNet introduced the Focal Loss to address the class imbalance in one-stage detectors, enhancing performance without altering the underlying network structure. FCOS redefined object detection by offering an anchor-free and proposal-free approach, operating pixel-by-pixel for a streamlined, full convolutional network-based solution. TOOD proposed a task-aligned one-stage detection method with a novel head structure and learning method, focusing on improving detection for specific tasks. VFNet designed the Varifocal Loss function to predict IoU-aware classification scores and introduced a star-shaped bounding box feature representation for refining object localization. YOLOv5 was a one-stage object detection algorithm, improved upon its predecessors by offering superior speed and accuracy in one-stage object detection.

\textit{Details:} All experiments are conducted using MMDetection\footnote{https://github.com/open-mmlab/mmdetection}, with dataset partitions following the default configuration. The experiments were optimized using the SGD optimizer. The momentum and weight decay are set to 0.9 and 0.0001, respectively. During the training and testing process, the image size was adjusted to 1333 * 800. The batch size is set to 2 for a single GPU, and distributed training is conducted using 4 GPUs. All models are trained on NVIDIA GeForce RTX 3090 GPUs, each with 24GB of graphics memory. We evaluate the mean average precision (mAP) of the object detection to measure the performance of all models fairly. The mAP(0.50:0.95) metric measures the mAP across different Intersection over Union (IoU) thresholds, ranging from 0.50 to 0.95 with an interval of 0.05. Similarly, mAP(0.50) is the mAP at a single IoU threshold of 0.50.

\subsubsection{Experimental Results}
\begin{table*}[htbp]
	\centering
	\setlength{\tabcolsep}{4pt}
	\caption{Comparisons of different methods on the Pictor-v3, SHWD, and SFCHD datasets}
	\label{tab:result_dataset}
	% \begin{threeparttable}
	\begin{tabular}{cccccccc}
	\toprule
	\multirow{2}{*}{Method}  &\multirow{2}{*}{Backbone} & \multicolumn{2}{c}{Pictor-v3} &\multicolumn{2}{c}{SHWD} & \multicolumn{2}{c}{SFCHD (ours)}\\
	\cmidrule(lr){3-4}  \cmidrule(lr){5-6}  \cmidrule(lr){7-8}
	 &\multirow{2}{*}{} & mAP(0.50) &mAP(0.50:0.95) &mAP(0.50) & mAP(0.50:0.95) &mAP(0.50) &mAP(0.50:0.95) \\
	\midrule
	SSD &VGG16 &0.855 &0.488 & 0.808 &0.574 &0.728 &0.415 \\
	Faster RCNN& ResNet-50 &0.906 &0.534 &0.848 &0.631 &0.764 &0.503 \\
	FCOS  &ResNet-50 & 0.895 &0.524 &0.858 &0.639 &0.764 &0.496 \\
	VFNet  &ResNet-50 &0.914 &0.552 &0.857 &0.639 & 0.764 &0.510 \\
	RetinaNet  &ResNet-50 &0.905 &0.544 &0.855 &0.636 &0.759 &0.489 \\
	TOOD  &ResNet-50 &0.915 &0.558 & 0.867 & 0.644 &0.789 & 0.523 \\
	YOLOv5  &CSPDarknet53 &0.882 &0.536 &0.840 &0.639 &0.741 &0.496 \\
	\bottomrule
	\end{tabular}
	% \end{threeparttable}
\end{table*}
Table~\ref{tab:result_dataset} presents a comparative analysis of various methods across different datasets. We primarily rely on two metrics, i.e., mAP(0.50) and mAP(0.50:0.95), to gauge the adaptability of different networks to each dataset. It is observable from the table that our SFCHD records the lowest scores in mAP, followed by the Pictor-V3 and SHWD datasets. This is primarily attributed to the higher complexity of our dataset, which encompasses intricate backgrounds, diverse categories, and undesirable lighting conditions, thereby presenting greater challenges. Undoubtedly, models trained on datasets that more closely mimic real-world scenarios and are more challenging tend to exhibit more effective performance in practical applications.

Table~\ref{tab:ap_category_dataset} delineates the performance of distinct categories in our dataset. It is observable from the table that YOLOv5 achieves a heightened level of precision in the identification of the Person, Safety Helmet, and Safety Clothing categories. This elevated accuracy may be attributed to the abundant data representation for these classes, thereby equipping the network with the capacity to absorb a more extensive array of characteristics. In the case of the Other Clothing and Head categories, despite their limited proportion within the dataset, the pronounced divergence in their features from the Safety Clothing and Safety Helmet facilitates the network's capability to distinguish and assimilate these variations. In contrast, the Blurred Clothing and Blurred Head categories, constrained by a scarcity of samples in the dataset and further impeded by complex environmental factors such as shadows and intense lighting, present a formidable challenge for the network to discern efficacious features, thereby leading to diminished discernment efficacy.
\begin{table*}[htbp]
	\centering
	\setlength{\tabcolsep}{5.5pt}
	\caption{Performance of different categories in the SFCHD dataset}
	\label{tab:ap_category_dataset}
	\begin{tabular}{ccccccccc}
		\toprule
		\multirow{2}{*}{Method} & Person & Safety Helmet & Safety Clothing & Other Clothing & Head & Blurred Clothing & Blurred Head \\
		\cmidrule(lr){2-8}
		&\multicolumn{7}{c}{AP(0.50:0.95)}& \\
		\midrule
		SSD & 0.602 & 0.565 & 0.557 & 0.452 & 0.384 & 0.205 & 0.142 \\
		Faster RCNN & 0.712 & 0.647 & 0.646 & 0.545 & 0.493 & 0.272 & 0.205 \\
		FCOS  & 0.684 & 0.634 & 0.646 & 0.540 & 0.480 & 0.289 & 0.196 \\
		VFNet  & 0.731 & 0.662 & 0.643 & 0.540 & 0.525 & 0.251 & 0.220 \\
		RetinaNet & 0.711 & 0.635 & 0.645 & 0.518 & 0.485 & 0.271 & 0.159 \\
		TOOD & 0.729 & 0.660 & 0.659 & 0.562 & 0.528 & 0.296 & 0.223 \\
		YOLOv5 & 0.727 & 0.664 & 0.637 & 0.549 & 0.507 & 0.212 & 0.189 \\
		\bottomrule
	\end{tabular}
\end{table*}

In Table~\ref{tab:map_fps_dataset}, we compare the mAP(0.50:0.95) and inference speed of various models on the SFCHD dataset. It can be observed that TOOD achieves the highest mAP, but its FPS is only 17.0; SSD attains the highest FPS, but its mAP is lower than all other models, with a score of only 0.4150. Upon comprehensive consideration, YOLOv5 realizes the optimal balance between accuracy and speed.
\begin{table}[htbp]
	\centering
	\setlength{\tabcolsep}{6pt}
	\caption{Results of precision and speed for different methods in the SFCHD dataset}
	\label{tab:map_fps_dataset}
	% \begin{threeparttable}
	\begin{tabular}{cccccccc}
	\toprule
	Method  & Backbone & mAP(0.50:0.95)  & FPS \\
	\midrule
	SSD & VGG16 & 0.415 & 57.5 \\
	Faster  RCNN  & ResNet-50 & 0.503 & 25.9 \\
	FCOS & ResNet-50 & 0.496 & 27.5 \\
	VFNet & ResNet-50 & 0.510 & 11.7 \\
	RetinaNet & ResNet-50 & 0.489 & 27.0 \\
	TOOD & ResNet-50 & 0.523 & 17.0 \\
	YOLOv5 & CSPDarknet53  & 0.496  & 46.3\\
	\bottomrule
	\end{tabular}
	% \end{threeparttable}
\end{table}

\subsection{Validity of the SCALE Module}
In this subsection, we integrate the SCALE module with detection algorithms and experiment on the public low-light dataset to validate the efficacy of SCALE.

\subsubsection{Experimental Setup}
\textit{Dataset:} To substantiate the efficacy of the SCALE module, we conduct extensive experiments on the ExDark~\cite{LOH201930} dataset. The dataset is a benchmark specifically designed to assess object detection algorithms under low-light conditions. It comprises a total of 7,363 low-light images, categorized into 10 varying illumination conditions, ranging from extremely dim to moderately low light, across 12 classes, with a total of 23,710 annotated instances. The dataset is partitioned into a training set with 5,896 images and a testing set with 1,467 images. The images within the dataset encompass a multitude of scenarios from indoor to outdoor settings and from urban to natural landscapes, ensuring the diversity and applicability of the dataset.

\textit{Methods:} The models we select primarily encompass three categories. The first category consists of low-light image enhancement methods, including KinD~\cite{zhang2019Kind}, MBLLEN \cite{Lv2018MBLLENLI}, and Zero-DCE~\cite{zero_DCE}. KinD, inspired by Retinex theory, decomposed the image into two components, one responsible for light regulation (illumination) and the other for degradation elimination (reflectance). MBLLEN extracted rich features at different levels, enhanced them through multiple sub-networks, and ultimately produced the output image through multi-branch fusion. Zero-DCE formulated the task of light enhancement as a specific curve estimation for images with a deep network, training a lightweight deep network to estimate the dynamic range adjustment of pixels and higher-order curves for a given image. The second category is low-light image object detection methods, including MAET~\cite{Cui_2021_ICCV_MAET}, DENet~\cite{Qin_2022_ACCV_DENet}, IAT-YOLO~\cite{Cui_2022_BMVC_IAT}, and PE-YOLO~\cite{10.1007/978-3-031-44195-0_14}. MAET, in a self-supervised manner, considered the physical noise model and image signal processing to learn the intrinsic visual structure by encoding and decoding real-world illumination degradation transformations. DENet enhanced the model's performance under adverse weather conditions by using a Laplacian pyramid to decompose each input image into a low-frequency component and several high-frequency components. IAT-YOLO proposed a lightweight and fast illumination adaptive transformer for normalizing sRGB images under low-light or underexposed/overexposed conditions. PE-YOLO introduced a detail processing module composed of a context branch and an edge branch to enhance image details, as well as a low-frequency enhancement filter to obtain low-frequency semantic information and avoid high-frequency noise. The third category consists of classical one-stage object detection methods, including FCOS~\cite{tianfcos_34}, VFNet~\cite{zhang36}, and TOOD~\cite{fengtood37}. These methods are primarily utilized for cascading with the SCALE module to form new end-to-end models.

\textit{Details:} The inaugural category of models is trained on the LOL~\cite{Chen2018Retinex} dataset, subsequently employing the trained models to process the ExDark dataset, thereby generating enhanced and brightened imagery. These images are then subsequently input into the YOLOv3~\cite{redmon_v3_31} detector to procure the detection outcomes of the low-light image enhancement techniques. In contrast, the second category of models directly integrates low-light images into the network for training, culminating in the ultimate detection results. To ensure a fair comparative analysis, the detector employed across both categories of methods is the YOLOv3. Throughout the training and testing phases, the dimensions of the images are uniformly adjusted to 608 * 608. The experimental outcomes are assessed using the average precision (AP) and mean average precision (mAP) as the definitive evaluative metrics, with the IoU threshold established at 0.5. For training, the batch size for an individual GPU is configured to 4, and distributed training is executed across 4 GPUs. The optimization algorithm utilized during training is SGD, with the learning rate set to 0.001 and the weight decay factor designated as 0.0005. The third category of methods is to validate the enhancement effect of the SCALE module under low-light conditions. We integrate the SCALE module with existing object detection algorithms to obtain the detector's output results. In this category, experiments are conducted based on the MMDetection framework. During training and testing, the size of the images is uniformly adjusted to 1333 * 800, with the IoU threshold set to 0.5, and AP and mAP are employed as the evaluative metrics. The batch size for a single GPU is set to 4, and distributed training is performed using the same 4 GPUs. The optimizer for training is the SGD, with the learning rate set to 0.01 and the weight decay factor set to 0.0001.

\subsubsection{Experimental Results}
\begin{table*}[htbp]
	\centering
	\setlength{\tabcolsep}{3.5pt}
	\caption{Performance comparisons between our SCALE-YOLO and existing models on the ExDark dataset}
	\label{tab:result_comparison_module}
  \begin{threeparttable}
	\begin{tabular}{cccccccccccccc}
		\toprule
		\multirow{2}{*}{Method} & Bicycle & Boat &  Bottle & Bus & Car & Cat & Chair & Cup& Dog & Motorbike & People & Table & \multirow{2}{*}{mAP (\%)}\\
		\cmidrule(lr){2-13}
		&\multicolumn{11}{c}{AP (\%)}& &\\
		\midrule
		YOLOv3 & 79.8 & 75.3  & 78.1 & 92.3 & 83.0 & 68.0 & 69.0 & 79.0  & 78.0  & 77.3  & 81.5  & 55.5  & 76.4\\
		KinD & 80.1 & 77.7  & 77.2 & 93.8 & 83.9 & 66.9 & 68.7 & 77.4  & 79.3  & 75.3  & 80.9  & 53.8  & 76.3\\
		MBLLEN & 82.0 & 77.3  & 76 .5 & 91.3 & 84.0 & 67.6 & 69.1 & 77.6  & 80.4  & 75.6  & 81.9  & \textbf{58.6}  & 76.8\\
		Zero-DCE & 84.1 & 77.6  & 78.3 & 93.1 & 83.7 & 70.3 & 69.8 & 77.6  & 77.4  & 76.3  & 81.0  & 53.6  & 76.9 \\
		\midrule
		MAET & 83.1 & 78.5  & 75.6 & 92.9 & 83.1 & 73.4 & 71.3 & 79.0  & 79.8  & 77.2  & 81.1  & 57.0  & 77.7 \\
		DENet & 80.4 & \textbf{79.7}  & 77.9 & 91.2 & 82.7 & 72.8 & 69.9 & 80.1  & 77.2  & 76.7  & 82.0  & 57.2  & 77.3\\
		IAT-YOLO & 79.8 & 76.9  & 78.6 & 92.5 & 83.8 & 73.6 & 72.4 & 78.6  & 79.0  & \textbf{79.0} & 81.1   & 57.7  & 77.8\\
		PE-YOLO & \textbf{84.7} & 79.2 & \textbf{79.3} & 92.5 & 83.9 & 71.5 & 71.7 & 79.7  & 79.7  & 77.3  & 81.8  & 55.3  & 78.0 \\
		\midrule
		SCALE-YOLO (Ours) & 81.3 & 79.3  & 78.2 & \textbf{93.9} & \textbf{84.2} & \textbf{75.5} & \textbf{74.9} & \textbf{82.3}  & \textbf{81.0}  & 77.5  & \textbf{82.5}  & 57.3  & \textbf{79.0}\\
		\bottomrule
	\end{tabular}
  These models include both low-light image enhancement methods and low-light image object detection methods.
  \end{threeparttable}
\end{table*}
\begin{table*}[htbp]
	\centering
	\setlength{\tabcolsep}{4.5pt}
	\caption{Performance improvements of the SCALE module on the ExDark dataset}
	\label{tab:result_improvement_module}
	\begin{tabular}{cccccccccccccc}
		\toprule
		\multirow{2}{*}{Method} & Bicycle & Boat &  Bottle & Bus & Car & Cat & Chair & Cup& Dog & Motorbike & People & Table & \multirow{2}{*}{mAP (\%)}\\
		\cmidrule(lr){2-13}
		&\multicolumn{11}{c}{AP (\%)}& &\\
		\midrule
		FCOS & 75.5 & 64.4 & 68.0 & 86.8 & 78.5 & 69.3 & 55.4 & 71.7 & 70.0 & 64.8 & 72.3 & 46.7 & 68.6 \\
		FCOS+SCALE & 75.1 & 66.6 & 73.5 & 89.9 & 78.9 & 67.0 & 57.2 & 72.8 & 74.2 & 67.3 & 72.0 & 45.9 & \textbf{70.0}\\
		\midrule
		VFNet & 77.4 & 70.5 & 76.6 & 90.6 & 81.8 & 67.1 & 59.4 & 71.8 & 72.6 & 70.6 & 77.7 & 53.3 & 72.5 \\
		VFNet+SCALE & 79.4 & 70.2 & 76.5 & 89.7 & 81.7 & 71.9 & 60.9 & 71.5  & 75.0 & 71.2 & 77.4 & 55.5 & \textbf{73.4} \\
		\midrule
		TOOD & 77.0 & 69.2 & 72.2 & 90.0 & 80.0 & 72.6 & 63.0 & 71.8 & 71.0 & 71.9 & 76.2 & 52.2 & 72.3 \\
		TOOD+SCALE & 77.9 & 70.0 & 78.3 & 90.0 & 80.7 & 69.1 & 62.0 & 72.4 & 73.7 & 69.2 & 78.1 & 54.2 & \textbf{73.0} \\
		\bottomrule
	\end{tabular}
\end{table*}
In Table~\ref{tab:result_comparison_module}, we report the performance comparison between the proposed SCALE-YOLO and existing models, encompassing both low-light image enhancement methods and low-light image object detection methods. Compared to YOLOv3, SCALE-YOLO demonstrates a 2.6\% improvement in mAP. KinD, MBLLEN, and Zero-DCE are image enhancement models aimed at improving image details by enhancing brightness, but they also tend to amplify image noise, affecting feature extraction and representation. Compared to these methods, SCALE-YOLO achieves better detection performance with improvements of 2.7\%, 2.2\%, and 2.1\%, respectively. From the table, it can be discerned that compared to image enhancement methods, object detection methods exhibit superior performance. This indicates that the enhancement networks' comprehension of images is more effective than simply increasing image contrast and brightness. In comparison to MAET, SCALE-YOLO achieves a 1.3\% increase in mAP, against DENet, the mAP improvement is 1.7\%, compared to IAT-YOLO, the mAP enhancement is 1.2\%, and when juxtaposed with PE-YOLO, the mAP increase amounts to 1\%. SCALE-YOLO employs spatial attention to learn the correlation of different regions in low-light images and channel attention to learn the correlation of different channels, thereby focusing more on important feature channels. 

Table~\ref{tab:result_improvement_module} reveals the performance enhancements achieved by integrating the SCALE module into existing single-stage object detection algorithms. The results indicate that on the ExDark dataset, the performance of all three detectors is significantly bolstered. Specifically, the FCOS+SCALE version, which integrates the SCALE module, improves the mAP by 1.4\% compared to the original FCOS; the VFNet+SCALE version, with the addition of the SCALE module, sees a 0.9\% increase in mAP; and TOOD+SCALE enhances performance by 0.7\% over the original TOOD. These outcomes demonstrate that the SCALE module effectively augments these detectors' ability to comprehend and recognize images under low-light conditions, thereby elevating detection capabilities.

\begin{table}[htbp]
	\centering
	\setlength{\tabcolsep}{10pt}
	\caption{Ablation analysis for different pathways in our SCALE module}
	\label{tab:ablation_module}
		\begin{tabular}{cccc}
			\toprule
			Method  & SAP & CAP & mAP (\%) \\
			\midrule
			YOLOv3 & -- & -- & 76.4 \\
			\midrule
			\multirow{3}{*}{SCALE-YOLO (Ours)} & \ding{51} & \ding{55} & 77.3 \\
			& \ding{55} & \ding{51} & 77.8 \\
			& \ding{51} & \ding{51}& \textbf{79.0} \\
			\bottomrule
		\end{tabular}
\end{table}
To validate the efficacy of the spatial and channel attention pathways within the SCALE module, we conduct ablation studies on these two components, as reflected in Table~\ref{tab:ablation_module}. The experimental results indicate that when only the SAP is activated, the model's mAP increased by 0.9\% compared to YOLOv3, confirming that SAP can effectively focus on key information in low-light images. When only the CAP is enabled, the mAP increased by 1.8\%, demonstrating that CAP aids the model in identifying and enhancing channels carrying significant information. By comparing the performance improvements of the two, we can infer that inter-channel relationships may be weighted more heavily than spatial location information. When both SAP and CAP are employed jointly, the model's mAP is further enhanced by 2.6\%. The aforementioned results substantiate that the SCALE module can generate more discriminative feature representations, thereby facilitating the detector's profound understanding of images in low-light environments.

\subsection{Application of SCALE on SFCHD Dataset}
In this subsection, we apply the SCALE module to the SFCHD dataset to demonstrate that SCALE can effectively enhance the detection quality on the SFCHD dataset.

\subsubsection{Experimental Setup}
To verify the performance of the SCALE module on the SFCHD dataset, we select a range of object detection models for testing, including FCOS~\cite{tianfcos_34}, VFNet~\cite{zhang36}, TOOD~\cite{fengtood37}, and YOLOv8\footnote{https://github.com/ultralytics/ultralytics}. We integrate the proposed SCALE module into these detectors to construct an end-to-end low-light object detection model, which is optimized solely by a single loss function. In terms of experimental setup, we continue with the configuration from previous experiments, utilizing distributed training across 4 GPUs and employing the SGD optimizer. The learning rate is uniformly set to 0.01, and the selected evaluation metrics are AP and mAP. For the FCOS, VFNet, and TOOD models, the input image size is adjusted to 1333 * 800 pixels, the IoU threshold is set to 0.5, the Batch Size per GPU is 2, and the weight decay factor is 0.0001. For YOLOv8, the image size is adjusted to 640 * 640 pixels, the IoU threshold is increased to 0.7, the Batch Size per GPU is increased to 16, and the weight decay factor is set to 0.0005.

\subsubsection{Experimental Results}
\begin{table*}[htbp]
	\centering
	\setlength{\tabcolsep}{2.0pt}
	\caption{Performance improvements of the SCALE module on the SFCHD dataset}
	\label{tab:result_improvement_SFCHD}
	\begin{tabular}{cccccccccc}
		\toprule
		\multirow{2}{*}{Method} & Person & Safety Helmet & Safety Clothing & Other Clothing & Head & Blurred Clothing & Blurred Head & \multirow{2}{*}{mAP (\%)}\\
		\cmidrule(lr){2-8}
		&\multicolumn{7}{c}{AP (\%)}& &\\
		\midrule
		FCOS & 68.4 & 63.4 & 64.6 & 54.0 & 48.0 & 28.9 & 19.6 & \textbf{49.6} \\
		FCOS+SCALE & 68.5 & 63.8 & 64.7 & 53.0 & 47.6 & 28.4 & 20.5 & 49.5 \\
		\midrule
		VFNet & 73.1 & 66.2 & 64.3 & 54.0 & 52.5 & 25.1 & 22.0 & 51.0 \\
		VFNet+SCALE & 73.2 & 66.5 & 64.4 & 53.9 & 52.1 & 25.8 & 23.7 & \textbf{51.4} \\
		\midrule
		TOOD & 72.9 & 66.0 & 65.9 & 56.2 & 52.8 & 29.6 & 22.3 & \textbf{52.3} \\
		TOOD+SCALE & 72.9 & 66.2 & 66.2 & 56.2 & 51.6 & 29.6 & 23.5 & \textbf{52.3} \\
		\midrule
		YOLOv8 & 73.6 & 64.9 & 67.5 & 58.5 & 45.5 & 32.3 & 23.9 & 52.2 \\
		YOLOv8+SCALE & 74.4 & 66.1 & 68.8 & 58.4 & 47.5 & 33.2 & 25.2 & \textbf{53.3} \\
		\bottomrule
	\end{tabular}
\end{table*}
Table 10 provides a detailed exposition of the enhancement effects of the SCALE module on object detection performance within the SFCHD dataset. The experimental outcomes demonstrate that the integration of our SCALE module contributes to the improvement of detection accuracy for the VFNet and YOLOv8 detection algorithms. Specifically, the incorporation of the SCALE module into VFNet resulted in a 0.4\% increase in mAP compared to the original model; and the amalgamation of the SCALE module with YOLOv8 realized a 1.1\% increment in mAP. These data robustly confirm the capability of the SCALE module in augmenting the detector's comprehension of low-light image information. However, on the FCOS and TOOD models, the introduction of the SCALE module did not lead to a significant performance improvement. An analysis of the reasons suggests that this may be due to the fact that, in addition to extreme lighting conditions, images in the SFCHD dataset are also accompanied by a plethora of other types of noise. The confluence of these factors may pose a considerable challenge to the FCOS and TOOD models, impacting the further enhancement of model performance. Therefore, for datasets that encompass extreme complexity, such as SFCHD, the SCALE module, when combined with high-performing detectors, can better leverage its advantages in low-light enhancement, showcasing its potential and value in the field of object detection.

\section{Conclusion}\label{conclusion}
In this work, we focus on the complex environments of real-world construction sites, proposing an industrial dataset and a low-light image enhancement module to advance artificial intelligence research in high-risk construction sites and its application in practical scenarios. Specifically, we have constructed a large-scale, highly complex, and realistic dataset called SFCHD for safety helmet and clothing detection. This dataset encompasses 12,373 images across 40 different scenarios, covering 7 categories with a total of 50,552 annotated instances. These images are sourced from authentic construction sites, ensuring the practical applicability of the data. To validate the usability of the SFCHD dataset, we experimented with various classic object detection algorithms and compared them with other relevant datasets. Moreover, addressing the performance limitations of existing algorithms under low-light conditions, we designed a plug-and-play low-light enhancement module named SCALE. This module extracts valid information from low-light images in both spatial and channel dimensions, significantly enhancing the performance of object detection models. Experimental results on the ExDarK and SFCHD datasets demonstrate that the introduction of the SCALE module has led to substantial improvements in the performance of object detectors under low-light conditions. 

Looking forward, we plan to apply the SFCHD dataset to a broader range of computer vision tasks, such as instance segmentation and image classification, to further expand its application scope and impact. Concurrently, we will continue to explore and propose more innovative and advantageous methods to effectively address challenges like complex background noise in industrial datasets, providing stronger support for computer vision research in industrial settings.

\bibliographystyle{IEEEtran}
\bibliography{IEEEabrv,sfchd-scale.bib}

\begin{thebibliography}{10}
\providecommand{\url}[1]{#1}
\csname url@samestyle\endcsname
\providecommand{\newblock}{\relax}
\providecommand{\bibinfo}[2]{#2}
\providecommand{\BIBentrySTDinterwordspacing}{\spaceskip=0pt\relax}
\providecommand{\BIBentryALTinterwordstretchfactor}{4}
\providecommand{\BIBentryALTinterwordspacing}{\spaceskip=\fontdimen2\font plus
\BIBentryALTinterwordstretchfactor\fontdimen3\font minus
  \fontdimen4\font\relax}
\providecommand{\BIBforeignlanguage}[2]{{%
\expandafter\ifx\csname l@#1\endcsname\relax
\typeout{** WARNING: IEEEtran.bst: No hyphenation pattern has been}%
\typeout{** loaded for the language `#1'. Using the pattern for}%
\typeout{** the default language instead.}%
\else
\language=\csname l@#1\endcsname
\fi
#2}}
\providecommand{\BIBdecl}{\relax}
\BIBdecl

\bibitem{Kelm-sensor1}
A.~Kelm, L.~Laußat, A.~Meins-Becker, D.~Platz, M.~J. Khazaee, A.~M. Costin,
  M.~Helmus, and J.~Teizer, ``Mobile passive radio frequency identification
  ({RFID}) portal for automated and rapid control of personal protective
  equipment ({PPE}) on construction sites,'' \emph{Automation in Construction},
  vol.~36, pp. 38--52, 2013.

\bibitem{Naticchia-sensor3}
B.~Naticchia, M.~Vaccarini, and A.~Carbonari, ``A monitoring system for
  real-time interference control on large construction sites,''
  \emph{Automation in Construction}, vol.~29, pp. 148--160, 2013.

\bibitem{Hua17}
C.~Zhu, Y.~He, and M.~Savvides, ``Feature selective anchor-free module for
  single-shot object detection,'' in \emph{Proceedings of the IEEE/CVF
  Conference on Computer Vision and Pattern Recognition}, June 2019.

\bibitem{nah12}
S.~Nah, T.~Hyun~Kim, and K.~Mu~Lee, ``Deep multi-scale convolutional neural
  network for dynamic scene deblurring,'' in \emph{Proceedings of the IEEE
  Conference on Computer Vision and Pattern Recognition}, July 2017.

\bibitem{Lecun-4}
Y.~Lecun, L.~Bottou, Y.~Bengio, and P.~Haffner, ``Gradient-based learning
  applied to document recognition,'' \emph{Proceedings of the IEEE}, vol.~86,
  no.~11, pp. 2278--2324, 1998.

\bibitem{Kolar5}
Z.~Kolar, H.~Chen, and X.~Luo, ``Transfer learning and deep convolutional
  neural networks for safety guardrail detection in 2d images,''
  \emph{Automation in Construction}, vol.~89, pp. 58--70, 2018.

\bibitem{nath6}
N.~Nath, T.~Chaspari, and A.~Behzadan, ``A transfer learning method for deep
  neural network annotation of construction site imagery,'' in
  \emph{Proceedings of the 18th International Conference on Construction
  Applications of Virtual Reality}, 2018, pp. 1--10.

\bibitem{nath7}
N.~D. Nath, T.~Chaspari, and A.~H. Behzadan, ``Single- and multi-label
  classification of construction objects using deep transfer learning
  methods,'' \emph{J. Inf. Technol. Constr.}, vol.~24, pp. 511--526, 2019.

\bibitem{nath8}
N.~D. Nath, A.~H. Behzadan, and S.~G. Paal, ``Deep learning for site safety:
  Real-time detection of personal protective equipment,'' \emph{Automation in
  Construction}, vol. 112, p. 103085, 2020.

\bibitem{gochoo2021safety}
M.~Gochoo, ``Safety helmet wearing dataset,'' in \emph{Mendeley Data}, 2021.

\bibitem{hu2018squeeze}
J.~Hu, L.~Shen, and G.~Sun, ``Squeeze-and-excitation networks,'' in
  \emph{Proceedings of the IEEE Conference on Computer Vision and Pattern
  Recognition}, 2018, pp. 7132--7141.

\bibitem{everingham2010}
M.~Everingham, L.~Van~Gool, C.~K.~I. Williams, J.~Winn, and A.~Zisserman, ``The
  pascal visual object classes (voc) challenge,'' \emph{International Journal
  of Computer Vision}, vol.~88, no.~2, pp. 303--338, 2010.

\bibitem{everingham2015}
M.~Everingham, S.~M.~A. Eslami, L.~Van~Gool, C.~K.~I. Williams, J.~Winn, and
  A.~Zisserman, ``The pascal visual object classes challenge: A
  retrospective,'' \emph{International Journal of Computer Vision}, vol. 111,
  no.~1, pp. 98--136, 2015.

\bibitem{lin2014microsoft}
T.-Y. Lin, M.~Maire, S.~Belongie, J.~Hays, P.~Perona, D.~Ramanan, P.~Dollár,
  and C.~L. Zitnick, ``{Microsoft COCO}: Common objects in context,'' in
  \emph{Computer Vision - ECCV 2014}, 2014, pp. 740--755.

\bibitem{russakovsky2015}
O.~Russakovsky, J.~Deng, H.~Su, J.~Krause, S.~Satheesh, S.~Ma, Z.~Huang,
  A.~Karpathy, A.~Khosla, M.~Bernstein, A.~C. Berg, and L.~Fei-Fei, ``Imagenet
  large scale visual recognition challenge,'' \emph{International Journal of
  Computer Vision}, vol. 115, no.~3, pp. 211--252, 2015.

\bibitem{kuznetsova2020}
A.~Kuznetsova, H.~Rom, N.~Alldrin, J.~Uijlings, I.~Krasin, J.~Pont-Tuset,
  S.~Kamali, S.~Popov, M.~Malloci, A.~Kolesnikov, T.~Duerig, and V.~Ferrari,
  ``The open images dataset v4,'' \emph{International Journal of Computer
  Vision}, vol. 128, no.~7, pp. 1956--1981, 2020.

\bibitem{feng2022review}
D.~Feng, A.~Harakeh, S.~L. Waslander, and K.~Dietmayer, ``A review and
  comparative study on probabilistic object detection in autonomous driving,''
  \emph{IEEE Transactions on Intelligent Transportation Systems}, vol.~23,
  no.~8, pp. 9961--9980, 2022.

\bibitem{9933424}
S.~Wang, Y.~Sun, Z.~Wang, and M.~Liu, ``{ST-TrackNet}: A multiple-object
  tracking network using spatio-temporal information,'' \emph{IEEE Transactions
  on Automation Science and Engineering}, vol.~21, no.~1, pp. 284--295, 2024.

\bibitem{elakkiya2022cervical}
R.~Elakkiya, V.~Subramaniyaswamy, V.~Vijayakumar, and A.~Mahanti, ``Cervical
  cancer diagnostics healthcare system using hybrid object detection
  adversarial networks,'' \emph{IEEE Journal of Biomedical and Health
  Informatics}, vol.~26, no.~4, pp. 1464--1471, 2022.

\bibitem{li2023ga2mif}
J.~Li, X.~Wang, G.~Lv, and Z.~Zeng, ``{GA2MIF}: Graph and attention based
  two-stage multi-source information fusion for conversational emotion
  detection,'' \emph{IEEE Transactions on Affective Computing}, vol.~15, no.~1,
  pp. 130--143, 2024.

\bibitem{kim2021uncertainty}
J.~U. Kim, S.~Park, and Y.~M. Ro, ``Uncertainty-guided cross-modal learning for
  robust multispectral pedestrian detection,'' \emph{IEEE Transactions on
  Circuits and Systems for Video Technology}, vol.~32, no.~3, pp. 1510--1523,
  2022.

\bibitem{9612580}
S.~Du, P.~Cai, T.~Hu, and T.~Ikenaga, ``Automatic foreground detection at 784
  fps for ultra-high-speed human–machine interactions,'' \emph{IEEE
  Transactions on Automation Science and Engineering}, vol.~19, no.~4, pp.
  3587--3600, 2022.

\bibitem{dalal2005histograms}
N.~Dalal and B.~Triggs, ``Histograms of oriented gradients for human
  detection,'' in \emph{2005 IEEE Computer Society Conference on Computer
  Vision and Pattern Recognition}, vol.~1, 2005, pp. 886--893 vol. 1.

\bibitem{viola2001rapid}
P.~Viola and M.~Jones, ``Rapid object detection using a boosted cascade of
  simple features,'' in \emph{Proceedings of the 2001 IEEE Computer Society
  Conference on Computer Vision and Pattern Recognition}, vol.~1, 2001, pp.
  511--518.

\bibitem{krizhevsky19}
A.~Krizhevsky, I.~Sutskever, and G.~E. Hinton, ``{ImageNet} classification with
  deep convolutional neural networks,'' \emph{Communications of the ACM},
  vol.~60, no.~6, pp. 84--90, 2017.

\bibitem{Xie23}
S.~Xie, R.~Girshick, P.~Dollar, Z.~Tu, and K.~He, ``Aggregated residual
  transformations for deep neural networks,'' in \emph{Proceedings of the IEEE
  Conference on Computer Vision and Pattern Recognition}, July 2017.

\bibitem{sahil2022survey}
S.~S.~A. Zaidi, M.~S. Ansari, A.~Aslam, N.~Kanwal, M.~Asghar, and B.~Lee, ``A
  survey of modern deep learning based object detection models,'' \emph{Digital
  Signal Processing}, vol. 126, pp. 1--19, 2022.

\bibitem{girshick28}
R.~Girshick, J.~Donahue, T.~Darrell, and J.~Malik, ``Rich feature hierarchies
  for accurate object detection and semantic segmentation,'' in
  \emph{Proceedings of the IEEE Conference on Computer Vision and Pattern
  Recognition}, June 2014.

\bibitem{platt27}
J.~Platt, ``Sequential minimal optimization: A fast algorithm for training
  support vector machines,'' Microsoft, Tech. Rep. MSR-TR-98-14, April 1998.

\bibitem{Girshick25}
R.~Girshick, ``{Fast R-CNN},'' in \emph{Proceedings of the 2015 IEEE
  International Conference on Computer Vision}, December 2015.

\bibitem{ren26}
S.~Ren, K.~He, R.~Girshick, and J.~Sun, ``{Faster R-CNN}: Towards real-time
  object detection with region proposal networks,'' \emph{IEEE Transactions on
  Pattern Analysis and Machine Intelligence}, vol.~39, no.~6, pp. 1137--1149,
  2017.

\bibitem{redmon_v1_29}
J.~Redmon, S.~Divvala, R.~Girshick, and A.~Farhadi, ``You only look once:
  Unified, real-time object detection,'' in \emph{Proceedings of the IEEE
  Conference on Computer Vision and Pattern Recognition}, June 2016.

\bibitem{redmon_v2_30}
J.~Redmon and A.~Farhadi, ``Yolo9000: Better, faster, stronger,'' in
  \emph{Proceedings of the IEEE Conference on Computer Vision and Pattern
  Recognition}, July 2017.

\bibitem{redmon_v3_31}
------, ``{YOLOv3}: An incremental improvement,'' \emph{arXiv preprint
  arXiv:1804.02767}, 2018.

\bibitem{bochkovskiy_v4_32}
A.~Bochkovskiy, C.-Y. Wang, and H.-Y.~M. Liao, ``{YOLOv4}: Optimal speed and
  accuracy of object detection,'' \emph{ArXiv}, vol. abs/2004.10934, 2020.

\bibitem{liussd_33}
W.~Liu, D.~Anguelov, D.~Erhan, C.~Szegedy, S.~Reed, C.-Y. Fu, and A.~C. Berg,
  ``{SSD}: Single shot multibox detector,'' in \emph{Computer Vision -- ECCV
  2016}, 2016, pp. 21--37.

\bibitem{tianfcos_34}
Z.~Tian, C.~Shen, H.~Chen, and T.~He, ``Fcos: Fully convolutional one-stage
  object detection,'' in \emph{Proceedings of the IEEE/CVF International
  Conference on Computer Vision}, October 2019.

\bibitem{zhang2019Kind}
Y.~Zhang, J.~Zhang, and X.~Guo, ``Kindling the darkness: A practical low-light
  image enhancer,'' in \emph{Proceedings of the 27th ACM International
  Conference on Multimedia}, ser. MM '19.\hskip 1em plus 0.5em minus
  0.4em\relax New York, NY, USA: Association for Computing Machinery, 2019, pp.
  1632--1640.

\bibitem{Lv2018MBLLENLI}
F.~Lv, F.~Lu, J.~Wu, and C.~S. Lim, ``{MBLLEN}: Low-light image/video
  enhancement using cnns,'' in \emph{British Machine Vision Conference}, 2018.

\bibitem{IA_yolo}
W.~Liu, G.~Ren, R.~Yu, S.~Guo, J.~Zhu, and L.~Zhang, ``Image-adaptive yolo for
  object detection in adverse weather conditions,'' in \emph{AAAI Conference on
  Artificial Intelligence}, 2021.

\bibitem{Qin_2022_ACCV_DENet}
Q.~Qin, K.~Chang, M.~Huang, and G.~Li, ``{DENet}: Detection-driven enhancement
  network for object detection under adverse weather conditions,'' in
  \emph{Proceedings of the Asian Conference on Computer Vision (ACCV)},
  December 2022, pp. 2813--2829.

\bibitem{Cui_2022_BMVC_IAT}
Z.~Cui, K.~Li, L.~Gu, S.~Su, P.~Gao, Z.~Jiang, Y.~Qiao, and T.~Harada, ``You
  only need 90k parameters to adapt light: a light weight transformer for image
  enhancement and exposure correction,'' in \emph{Proceedings of the 33rd
  British Machine Vision Conference}, 2022.

\bibitem{linfocal38}
T.-Y. Lin, P.~Goyal, R.~Girshick, K.~He, and P.~Dollar, ``Focal loss for dense
  object detection,'' in \emph{Proceedings of the IEEE International Conference
  on Computer Vision}, Oct 2017, pp. 2980--2988.

\bibitem{fengtood37}
C.~Feng, Y.~Zhong, Y.~Gao, M.~R. Scott, and W.~Huang, ``{TOOD}: Task-aligned
  one-stage object detection,'' in \emph{Proceedings of the IEEE/CVF
  International Conference on Computer Vision}, October 2021, pp. 3490--3499.

\bibitem{zhang36}
H.~Zhang, Y.~Wang, F.~Dayoub, and N.~Sunderhauf, ``{VarifocalNet}: An iou-aware
  dense object detector,'' in \emph{Proceedings of the IEEE/CVF Conference on
  Computer Vision and Pattern Recognition}, June 2021, pp. 8514--8523.

\bibitem{LOH201930}
Y.~P. Loh and C.~S. Chan, ``Getting to know low-light images with the
  exclusively dark dataset,'' \emph{Computer Vision and Image Understanding},
  vol. 178, pp. 30--42, 2019.

\bibitem{zero_DCE}
C.~Guo, C.~Li, J.~Guo, C.~C. Loy, J.~Hou, S.~Kwong, and R.~Cong,
  ``Zero-reference deep curve estimation for low-light image enhancement,'' in
  \emph{2020 IEEE/CVF Conference on Computer Vision and Pattern Recognition
  (CVPR)}, 2020, pp. 1777--1786.

\bibitem{Cui_2021_ICCV_MAET}
Z.~Cui, G.-J. Qi, L.~Gu, S.~You, Z.~Zhang, and T.~Harada, ``Multitask {AET}
  with orthogonal tangent regularity for dark object detection,'' in
  \emph{Proceedings of the IEEE/CVF International Conference on Computer Vision
  (ICCV)}, October 2021, pp. 2553--2562.

\bibitem{10.1007/978-3-031-44195-0_14}
X.~Yin, Z.~Yu, Z.~Fei, W.~Lv, and X.~Gao, ``{PE-YOLO}: Pyramid enhancement
  network for dark object detection,'' in \emph{Artificial Neural Networks and
  Machine Learning -- ICANN 2023}, 2023, pp. 163--174.

\bibitem{Chen2018Retinex}
C.~Wei, W.~Wang, W.~Yang, and J.~Liu, ``Deep retinex decomposition for
  low-light enhancement,'' in \emph{Proceedings of the 29th British Machine
  Vision Conference}, 2018, pp. 1--12.

\end{thebibliography}
\balance
% \newpage

\end{document}